\documentclass[transmag]{IEEEtran}
\usepackage{graphicx}
\usepackage{amsfonts,amssymb,amsmath}
\usepackage{hyperref}
\usepackage[T1]{fontenc}
\usepackage{multirow}
\usepackage{adjustbox}
\usepackage{float}
\floatstyle{plaintop} 
\restylefloat{table}
\usepackage{txfonts}
\usepackage{cite}
\usepackage{textcomp}
\usepackage[style=base]{caption}
\usepackage{subcaption}
\usepackage{balance}
\usepackage{xcolor,cite,etoolbox}
\usepackage[labelformat=simple]{subcaption}

\def\BibTeX{{\rm B\kern-.05em{\sc i\kern-.025em b}\kern-.08em T\kern-.1667em\lower.7ex\hbox{E}\kern-.125emX}}
\begin{document}
\title{A Glimpse of Physical Layer Decision Mechanisms: Facts, Challenges, and Remedies}
\author{Selen Gecgel Cetin,~\IEEEmembership{Student Member,~IEEE},~Caner~Goztepe,\\Gunes~Karabulut~Kurt,~\IEEEmembership{Senior Member,~IEEE},~and~Halim Yanikomeroglu,~\IEEEmembership{Fellow,~IEEE} 
	\thanks{{S. Gecgel Cetin, and C. Goztepe are with the Department of Electronic and Communications Engineering, Istanbul Technical University, Istanbul, Turkey, 34469 (e-mails:\{gecgel16, goztepe\}@itu.edu.tr).}}
	\thanks{{G. Karabulut Kurt is with the Poly-Grames Research Center, Department of Electrical Engineering, Polytechnique Montreal, Montreal, QC, Canada, (e-mail: gunes.kurt@polymtl.ca).}}
	\thanks{{H. Yanikomeroglu is with the Department of Systems and Computer Engineering, Carleton University, Ottawa, ON K1S 5B6, Canada (e-mail: halim@sce.carleton.ca).}}}
\IEEEtitleabstractindextext{
\begin{abstract} Communications are realized as a result of successive decisions at the physical layer, from modulation selection to multi-antenna strategy, and each decision affects the performance of the communication systems. Future communication systems must include extensive capabilities as they will encompass a wide variety of devices and applications. Conventional physical layer decision mechanisms may not meet these requirements, as they are often based on impractical and oversimplifying assumptions that result in a trade-off between complexity and efficiency. By leveraging past experiences, learning-driven designs are promising solutions to present a resilient decision mechanism and enable rapid response even under exceptional circumstances. The corresponding design solutions should evolve following the lines of learning-driven paradigms that offer more autonomy and robustness. This evolution must take place by considering the facts of real-world systems and without restraining assumptions. In this paper, the common assumptions in the physical layer are presented to highlight their discrepancies with practical systems. As a solution, learning algorithms are examined by considering the implementation steps and challenges. Furthermore, these issues are discussed through a real-time case study using software-defined radio nodes to demonstrate the potential performance improvement. A cyber-physical framework is presented to incorporate future remedies.
\end{abstract}
\begin{IEEEkeywords}
Cybertwin, decision mechanisms, learning-driven solutions, machine learning, physical layer, real-world impairments.
\end{IEEEkeywords}}
\maketitle
\section{Introduction}\label{intro}
\IEEEPARstart{T}{owards} the sixth-generation (6G) networks, flexible and ubiquitous connectivity is expected, even under extraordinary conditions. Numerous technologies are envisioned to achieve this goal. On the other hand, heterogeneity of application domains for these technologies constitutes a significant need for more customized deployments. Therefore, stringent physical layer (PHY) requirements are emerging in terms of fast-responsiveness, reliability, latency, spectral efficiency, and security \cite{int1,int2,int3}.

\begin{figure}[t!]
	\centering
	\includegraphics[width=0.5\textwidth]{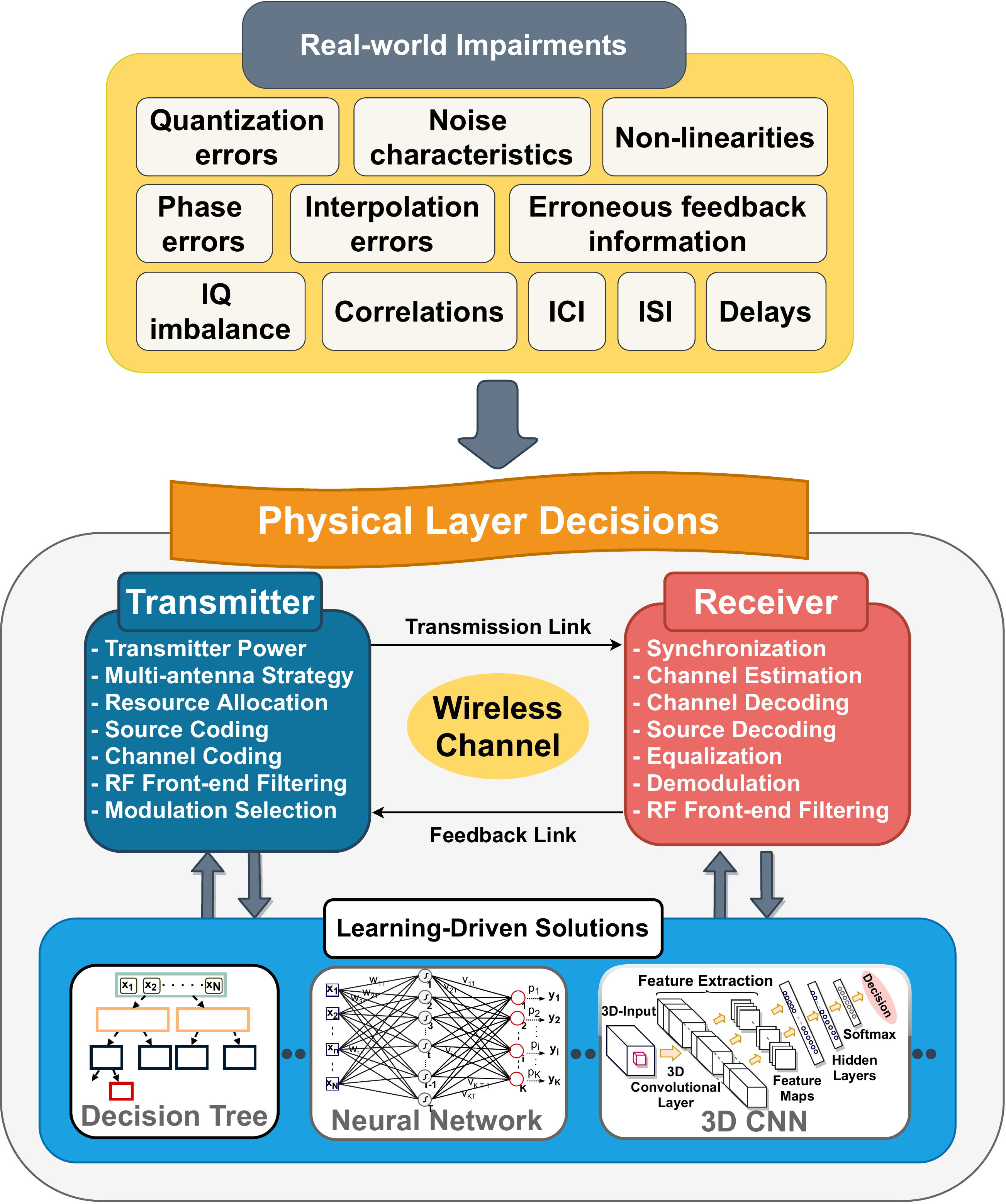}
	\caption{A demonstration of major decision steps in PHY layer and the interaction of learning-driven solutions.}
	\label{fig1}
\end{figure}

A typical digital communication system contains several signal processing blocks at the transmitter and receiver sides, such as equalization, bandpass signaling, channel coding, multiplexing and multiple access. Therefore, communication systems at the PHY can be defined as consecutive decision mechanisms that separately handle a subproblem and find appropriate decisions for signal processing blocks. Each decision, from transmit power to coding rate, jointly ensures efficient transmission of information bits. Although optimal decisions are theoretically attainable for the most part of problems, they may not be available under real-world impairments such as correlations or delays, as demonstrated in Fig. \ref{fig1}. Theoretical implications are generally based on assumptions that benefit from the advantages of simplicity, such as effortlessness to achieve an understandable formulation, perceptibility, and interpretability. They allow us to discover the theoretical bounds of the proposed system and give rise to diversify sub-optimal methods by compromising on a number of necessity. On the contrary, modeling mismatches due to these assumptions distract the solutions from actual systems, especially in real-world implementations. Current decision mechanisms for wireless communications are largely modeled under oversimplifying assumptions, as listed in Table \ref{assump}. However, future implementations will require re-evaluation of the solutions for PHY decision mechanisms to meet demanding targets.

Over the past decade, learning-driven approaches have been considered the leading candidates to achieve the ambitious goals of next-generation communication systems. Several studies are motivated with the advantages of learning-driven solutions, such as the elimination of human intervention and the use of big data stacks in PHY. In \cite{int7}, three machine learning algorithms are separately applied for channel assignment problem to evade the high complexity of the convex optimization based algorithm. The results show that the time complexity can be reduced without making concessions from the prediction accuracy. The study in \cite{int12} compares the performance of machine learning assisted and the conventional threshold-based link adaptation schemes. The learning-driven scheme outperforms its counterpart by achieving higher throughput. In \cite{int8}, a deep neural network is integrated into Viterbi algorithm instead of log-likelihood computations and it can be reliably employed in complex channel models thanks to its dynamic channel tracking capability. The results indicate the robustness, success, and adaptability of the learning-driven solutions. In \cite{int9}, power control, interference coordination, and beamforming are enabled by using deep reinforcement learning without the knowledge of channel state information (CSI). This study shows that learning algorithms can provide competent systems capable of making joint decisions with instantly responsive designs.

The aforementioned studies seem highly incentive for further researches, but offer no guarantee that the same results will be obtained under real conditions. The fact that learning-driven solutions do not outperform the optimal solution in the presence of ideal conditions should be kept in mind. However, instead of solutions based on impractical assumptions, they can provide solutions that go hand in hand with environmental changes and are based on real systems' attributes \cite{int5,int6,ref12,ref15,int11}. In \cite{int5}, neural network topologies are examined for non-coherent demodulation, which is convenient but difficult to model optimally for practical wireless communication systems due to non-linearities, non-stationarity, and non-Gaussian noise. The authors of \cite{int6} sign the complex channel conditions without a mathematically tractable model and design the communication system as an autoencoder. Adaptive transmission and generalized data representation schemes are proposed to maximize the data rate under different channel conditions. Block error rate and minimum mean squared error performance of the schemes are inspiring at lower signal-to-noise ratio (SNR). The study in \cite{ref12} compares learning-driven algorithms with the classical Euclidean distance-based method for the antenna selection problem considering channel imperfections and correlation. The results over a real-time test bed show that learning-driven decision mechanisms are also able to track correlation and deal with channel imperfections besides improving error performance. In \cite{ref15}, a PHY security problem is addressed, and neural networks acting as detector and identifier show high accuracy under actual conditions. The authors of \cite{int11} address the self-interference problem of flexible duplexing that take place in specifications. The study applies machine learning to the tuning process of radio frequency (RF) canceller and achieves the fastest convergence.

The accomplishments of above studies evince that learning algorithms are not only an alternative to their classical counterparts, but also very well suited to deal with PHY challenges. Most part of the studies analyze the deep-learning algorithms by presenting challenges and applications \cite{ref5, ref6, ref18}. In this paper, an emerging necessity in PHY, \textit{a changeover to learning-driven decision mechanisms} is highlighted. To this end, our major contributions are listed below:	
\begin{itemize}
\item Design issues for PHY decision mechanisms are addressed under five aspects: Synchronization, channel estimation errors, erroneous feedback information, RF front-end impairments, and correlation. The main incompatibilities between the real systems and the underlying assumptions (used in the derivations and simulations) of the current theoretical implications are outlined for each aspect in Table \ref{assump} (Section \ref{sectionII}).
\item We methodologically discuss the justifications to changeover from current methodologies to future methodologies and present the critical observations that must be taken into account for new solutions. The motivations why machine learning algorithms should be part of future decision mechanisms instead of some current approaches are addressed (Section \ref{sectionIII}).
\item We present an insightful roadmap that provides a perspective on the development of machine learning-driven PHY decision mechanisms for real communication systems (Section \ref{sectionIV}).
\item In the spirit of fair analysis, we also discuss the main challenges for learning-driven solutions (Section \ref{sectionV}).
\item The eligibility of learning-driven solutions for real systems is discussed through the problem of antenna selection over a test-bed of software defined radio (SDR) nodes. The discussions are extended for different analog-to-digital/digital-to-analog (ADC /DAC) resolutions (Section \ref{sectionVI}).
\item Four learning algorithms are selected by considering the time and computational complexities. Performances of decision tree (DTREE), multi-layer perceptron (MLP), random forest (RForest), and convolutional neural network (CNN) algorithms are investigated via the real-time SDR-based case study (Section \ref{sectionVI}).
\item We present an intelligent cyber-physical framework towards tight operational requirements of 6G in the presence of stringent PHY constraints. We provide insight into how learning-driven solutions interoperate with the other remedies thanks to the proposed framework (Section \ref{sectionVII}).
\end{itemize}
The remainder of this article is organized as follows. Section \ref{sectionII} addresses the design issues under five aspects related to PHY decision mechanisms for real systems. Section \ref{sectionIII} gives a methodological overview of current solutions and presents the reasons and motivations why future solutions should focus on machine learning algorithms. Section \ref{sectionIV} clarifies the roadmap for researchers to present more practical learning-driven solutions. Key challenges for learning-driven solutions are discussed in Section \ref{sectionV}. Section \ref{sectionVI} proposes four learning-driven solutions to the antenna selection problem and compares their performances with a classical approach by discussing the considerations from the previous sections. Section \ref{sectionVII} presents a novel framework that encompasses the 6G remedies and not only the learning-driven solutions. The paper concludes with Section \ref{sectionVIII}.
\begin{table*}[!ht]
	\centering
	\caption{An overview of the simplifying assumptions and challenging facts in PHY.}
	\label{assump}
	\begin{adjustbox}{max width=\textwidth}
		{
			
			\begin{tabular}{|c|c|l|l|} 
				\hline
				\begin{tabular}[c]{@{}c@{}} \textbf{\textcolor[rgb]{0.055,0.063,0.102}{Technical Aspects}}\\ \end{tabular}            & \textbf{Issues}                                                                               & \multicolumn{1}{c|}{\textbf{Assumptions} }                                                                                                                                                                                                                                                                                                            & \multicolumn{1}{c|}{\textbf{Facts} }                                                                                                                                                                                                                                                        \\ 
				\hline
				\multirow{4}{*}{\textbf{Synchronization} }                                                                            & \begin{tabular}[c]{@{}c@{}}\textit{Initial}\\\textit{Acquisition} \end{tabular}               & \begin{tabular}[c]{@{}l@{}}- Prior probabilities and pre-determined threshold are always valid\\for hypothesis tests. \end{tabular}                                                                                                                                                                                                                   & - False alarm can occur when there is no reference signal.                                                                                                                                                                                                                                  \\ 
				\cline{2-4}
				& \begin{tabular}[c]{@{}c@{}}\textit{Carrier}\\\textit{Phase} \end{tabular}                     & \begin{tabular}[c]{@{}l@{}}- No ISI.\\ - Channel provides the Nyquist criterion. \\ - Estimated phase variance satisfies Cramer-Rao bound (then optimal). \end{tabular}                                                                                                                                                                               & \begin{tabular}[c]{@{}l@{}}- ISI can be realized in practice: ISI suppression should be added to\\approximate the assumption. \end{tabular}                                                                                                                                                 \\ 
				\cline{2-4}
				& \begin{tabular}[c]{@{}c@{}}\textit{Carrier}\\\textit{Frequency}\\\textit{Offset}\end{tabular} & \begin{tabular}[c]{@{}l@{}}- The carrier phase is constant or does not change significantly. \\ - No Doppler shift. \\ - Channel satisfies Nyquist criteria. \end{tabular}                                                                                                                                                                            & - The carrier phase can change in time.                                                                                                                                                                                                                                                             \\ 
				\cline{2-4}
				& \begin{tabular}[c]{@{}c@{}}\textit{Timing}\\\textit{Offset}\end{tabular}                      & \begin{tabular}[c]{@{}l@{}}- No ISI. \\- No sampling delay. \\- Channel conditions are known. \end{tabular}                                                                                                                                                                                                                                           & \begin{tabular}[c]{@{}l@{}}- ISI generally exists.\\- CSI is estimated with errors due to TO.\\- Delays can not be eliminated entirely.\end{tabular}                                                                                                                                        \\ 
				\hline
				\multirow{2}{*}{\begin{tabular}[c]{@{}c@{}}\textbf{Channel}\\\textbf{Estimation}\\\textbf{Errors} \end{tabular}}      & \begin{tabular}[c]{@{}c@{}}\textit{Quantization}\\\textit{Errors} \end{tabular}               & \begin{tabular}[c]{@{}l@{}}- The optimum threshold to quantize is known. \\- The quantization error is uniformly distributed. \\- ADCs and DACs operate ideally and do not introduce any\\non-linearities for channel estimation and signal detection. \end{tabular}                                                                                  & \begin{tabular}[c]{@{}l@{}}- The converters in the transceiver may have different resolutions.\\- The threshold can change depending on the environmental conditions. \\- Quantization noise may have different statistical characteristics due to\\its time-varying nature. \end{tabular}  \\ 
				\cline{2-4}
				& \begin{tabular}[c]{@{}c@{}}\textit{Interpolation}\\\textit{Errors} \end{tabular}              & \begin{tabular}[c]{@{}l@{}}- Perfect knowledge of signal autocorrelation and noise\\variance is available (then Wiener interpolator is optimal).\end{tabular}                                                                                                                                                                                         & - Excellent knowledge about the signal and noise is not attainable.                                                                                                                                                                                                                         \\ 
				\hline
				\multirow{2}{*}{\begin{tabular}[c]{@{}c@{}}\textbf{Erroneous}\\\textbf{Feedback}\\\textbf{Information} \end{tabular}} & \begin{tabular}[c]{@{}c@{}}\textit{Feedback}\\\textit{Errors} \end{tabular}                   & \begin{tabular}[c]{@{}l@{}}- Reverse (feedback) channel is noise-free.\\- Feedback channel is not bandwidth limited.\\- Feedback overhead can be negligible.\\- Reverse link has the same statistical reciprocity. \\- Receiver knows perfectly the forward channel or has knowledge\\related to feedback errors. \end{tabular}                       & \begin{tabular}[c]{@{}l@{}}- The equalizer may be unstable.\\- The channel coefficients can change in time.\\- The channel coefficients can be correlated in time, frequency, and space.\\ \end{tabular}                                                                                            \\ 
				\cline{2-4}
				& \begin{tabular}[c]{@{}c@{}}\textit{Feedback}\\\textit{Delays} \end{tabular}                   & \begin{tabular}[c]{@{}l@{}}- There are no feedback delays.\\- The forward channel and feedback channel delays are known. \end{tabular}                                                                                                                                                                                                                & - The instantaneous feedback information is not available.                                                                                                                                                                                                                                  \\ 
				\hline
				\multirow{3}{*}{\begin{tabular}[c]{@{}c@{}}\textbf{RF}\\\textbf{Front-end}\\\textbf{Impairments} \end{tabular}}       & \begin{tabular}[c]{@{}c@{}}\textit{IQ}\\\textit{Imbalance} \end{tabular}                      & - No imbalance between I and Q.                                                                                                                                                                                                                                                                                                                       & \begin{tabular}[c]{@{}l@{}}- IQ imbalance leads to performance degradations due to phase\\rotations and signal distortions. \end{tabular}                                                                                                                                                   \\ 
				\cline{2-4}
				& \begin{tabular}[c]{@{}c@{}}\textit{Phase}\\\textit{Noise} \end{tabular}                       & \begin{tabular}[c]{@{}l@{}}- No frequency deviation at the output of the RF oscillator.\\- The higher carrier frequencies do not increase the phase noise. \end{tabular}                                                                                                                                                                              & \begin{tabular}[c]{@{}l@{}}- RF oscillators are low-cost devices and have instabilities.\\- Higher carrier frequencies lead to an increase in errors and residual terms. \end{tabular}                                                                                                      \\ 
				\cline{2-4}
				& \textit{Non-linearities}                                                                      & - The electronic units work ideally.                                                                                                                                                                                                                                                                                                                  & - The perfect stability is unrealistic for active devices.                                                                                                                                                                                                                                  \\ 
				\hline
				\multirow{3}{*}{\textbf{Correlation} }                                                                                & \begin{tabular}[c]{@{}c@{}}\textit{Time}\\\textit{Correlation} \end{tabular}                  & \begin{tabular}[c]{@{}l@{}}- Information sources are statistically stationary, wide-sense stationary, \\ergodic, or cyclostationary random processes.\\- The channel is merely time-invariant. \end{tabular}                                                                                                                                           & \begin{tabular}[c]{@{}l@{}}- The stationarity conditions can not always exist practically. \\- The channel characteristics can vary because of the external factors and\\mobility but show mnemonic relationships. \end{tabular}                                                            \\ 
				\cline{2-4}
				& \begin{tabular}[c]{@{}c@{}}\textit{Spatial}\\\textit{Correlation} \end{tabular}               & \begin{tabular}[c]{@{}l@{}}- The position of the transmitter and the receiver do not change. \\- The channel correlation matrix of the users is equal to zero.\\- If present, the correlation between MIMO channels is constant.\\- The receiver or transmitter moves slowly, and the received signals\\are not impressed by shadowing. \end{tabular} & \begin{tabular}[c]{@{}l@{}}- Each channel varies instantaneously.\\- User channels may not be statistically independent. \\- The current scenarios include the high mobility and the ones will be in 3D. \end{tabular}                                                                      \\ 
				\cline{2-4}
				& \begin{tabular}[c]{@{}c@{}}\textit{Frequency}\\\textit{Correlation} \end{tabular}             & \begin{tabular}[c]{@{}l@{}}- The frequency correlation of the channel is related to the\\power delay profile. \end{tabular}                                                                                                                                                                                                                          & - The scattering is not always wide-sense stationary and uncorrelated.                                                                                                                                                                                                                   \\
				\hline
			\end{tabular}

		}
	\end{adjustbox}
\end{table*}
\section{Five Design Aspects for PHY Decision Mechanisms}\label{sectionII}
{In this section, we present a projection of five compelling aspects to obtain superior PHY decisions considering real-world systems. Based on these aspects, simplifying assumptions and challenging facts of PHY are addressed and listed in Table \ref{assump}.}

\subsection{Synchronization}
Synchronization is the major indispensability for the compatibility of the transmitter's and receiver's processes and for accurate transmission \cite{ref7}. Focusing on PHY, the receiver must detect the suitable times to sample the transmitted signal and compensate the oscillators' phase and frequency errors depending on the received signal. A large number of studies assume that transmitter and receiver are perfectly synchronized and prove the success with numerical results. However, a residual synchronization error may remain in time and frequency due to the factual circumstances. Contrary to idealized systems, the real-world systems must include more qualifications due to the RF front-end impairments, mobility, variety of channel conditions, and delays \cite{S4,S11}. They can include several processes to reach a better synchronization, such as initial time synchronization between the transmitter and receiver, a robust lock mechanism, the estimation of carrier frequency offset (CFO) and timing offset (TO), algorithms for adaptable redundancy insert, and synchronization recovery. 

Estimation plays a critical role in reaching tenable synchronization. Theoretically optimal data-assisted methods are presented, including the hypothesis test for the initial acquisition, maximum-likelihood based estimation algorithms for CFO, TO, and phase offset \cite{S1,S2,S3,S4,S5,S6,S7,S9,S11,S12,S13,S14,S15,S16}. However, there may not be sufficient statistics for these methods without idealizing the systems. For example, the synchronization parameters may not be constant, or the variance of the synchronization error may not be exactly calculated due to random fluctuations caused by the phase noise of the oscillator. If we view other solutions, the majority of sub-optimal methods comprise moderately or based on the availability of the ideal CSI. In practice, propagation delay is generally unknown, intersymbol interference (ISI) or interchannel interference (ICI) may occur, and RF impairments may exist \cite{ref8,S8,S10}. Therefore, the existing methods should be evolved or new approaches should be improved considering the facts of the actual systems.

\subsection{Channel Estimation Errors}
Accurate channel estimation is critical to realize signal processing steps in PHY. The channel can be identified by correlation, maximum-likelihood, maximum a posteriori, or least-squares based estimators \cite{ref9,CE1,CE2,CE3}. These methods hinge on several assumptions, as listed in Table \ref{assump}, that are not always feasible from practical aspects. The acquisition of ideal CSI is not always available contrary to the common conjecture. For example, conventional systems assume that ADCs have infinite resolution, and there is no quantization error. The real systems have to overcome estimation errors that cause round-off errors, unavailability of instantaneous feedback information, and interpolation errors. On the other hand, the impact of estimation errors on performance will be more critical due to massive antennas, 3D deployments, and high mobility scenarios because the channel conditions in future networks will change rapidly, and channel estimation errors must be eliminated for a superior performance \cite{CE4,CE5,CE6}. Therefore, real-world impairments must be considered, and the solutions should be evaluated in terms of versatility.
\subsection{Erroneous Feedback Information}
Feedback information is the primary requirement to coordinate the transmitter and receiver. It affects the overall performance. However, it is mostly erroneous by virtue of over-optimistic assumptions, as summarized in Table \ref{assump}. The feedback information obtained via the forward link may not ideally represent the reverse link since they are not available simultaneously. Another issue is the outdated feedback caused by feedback delays \cite{F1,F2,F4,F5}. Conversely, it must include all variations to ideally adjust the processes, such as antenna selection or beamforming. Consequently, the erroneous and delayed feedback information leads to a decrease in overall system performance \cite{F3,F6}. To feed the entire system with more accurate information, we have to develop a more intelligent and cautious solutions.

\begin{figure*}[!ht]
	\centering
	\includegraphics[width=0.75\textwidth]{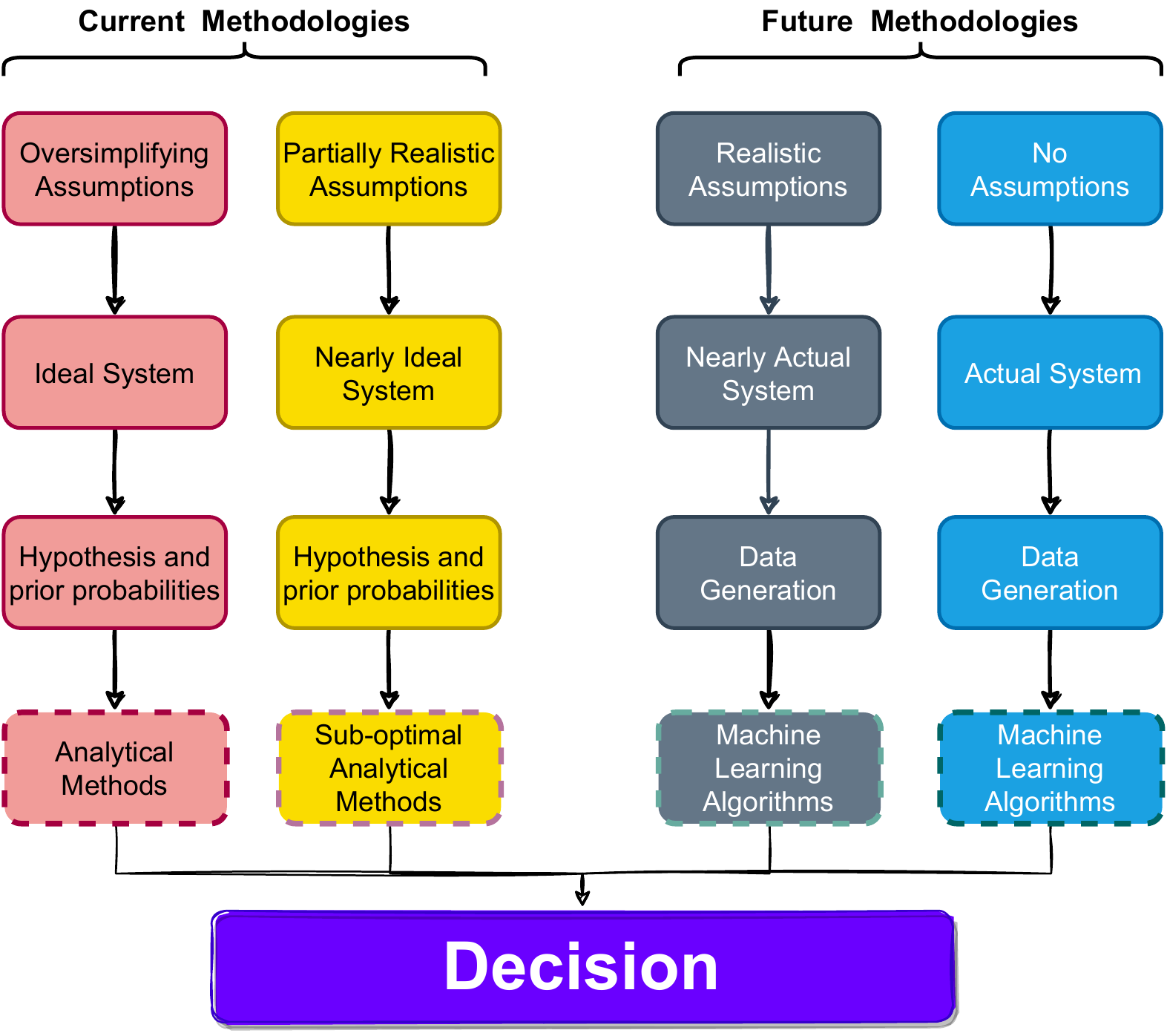}
	\caption{A comparative overview of current and future methodologies for PHY decision mechanisms.}
	\label{method}
\end{figure*}
\subsection{RF Front-end Impairments}
RF front-end impairments are apparent differences between real-world systems and simulated systems, and they are inevitable. Cost constraints and the fact that RF incompatibilities are hardware-based make them difficult to handle. Here, we addressed three main issues with respect to the assumptions in Table \ref{assump} that significantly affect overall system performance.
\subsubsection{In-phase/quadrature-phase (IQ) Imbalance}
IQ imbalance is the mismatch of the amplitude and/or phase between in-phase (I) and quadrature-phase (Q) components. It stems from low-cost devices and causes a performance degradation; therefore, it must be considered in real systems \cite{S11,IQ1,IQ2}. IQ imbalance restrains the systems by sensitizing towards other impairments such as CFO. Estimation algorithms to compensate IQ imbalance based on least mean squares, maximum-likelihood, expectation-maximization, or iterations exist, but they introduce extra computation.
\subsubsection{Phase Noise}
The phase noise in the oscillator results from the active circuit elements and makes frequency adjustments difficult. These fluctuations result in common phase error (CPE) and ICI. The real systems have to include a suppression process for ICI and an estimation algorithm for CPE mitigation \cite{ref10}.
\subsubsection{Non-linearities}
{Non-linearities are realized by ADC/DAC, mixers, and amplifiers (the power amplifier in the transmitter and the low-noise amplifier in the receiver). The main part of the non-linearities in PHY is caused by power amplifiers. They are mostly ignored due to the difficulties of theoretically modeling.}
\subsection{Correlation}
Formulation of correlation is not straightforward due to its uncertainty. Making assumptions may cause a fallacious representation of the correlation properties, leading to an inaccurate model. Furthermore, describing correlation can require continuous pattern tracing in time, frequency, or spatial domains, which introduces additional processes and increases complexity. For example, auto-correlation of information sequence in time and frequency domains or cross-correlations between different channels introduce the requirement of new computational blocks in the transmitter and receiver.
\section{Evaluations of PHY Decision Mechanisms}\label{sectionIII}
The key point to obtain the optimal decision in PHY is structuring the problem in a linear model by assuming the system is minimum variated and unbiased \cite{ref1}. Current methodologies {\textendash{ as represented in Figure \ref{method} }\textendash} propose analytical methods by following this principle. The reason is that simplifying assumptions provide a reasonable and effortless means to examine the systems that are challenging to model. The majority of the solutions introduce oversimplifying assumptions {\textendash{ as listed in Table \ref{assump} }\textendash} and idealize the system. Although some studies construct their solutions with partially realistic assumptions, they must make concessions in optimality or increase the complexity of the solution. After assuming that the system is ideal or nearly ideal, they define the problem with simplified possible conditions to constitute analytical methods. Consequently, the solution modeled by the current methodologies gradually deviates from the actual system step-by-step.
\subsection{Critical Observations}
Future methodologies must observe communication systems as in actual systems to make PHY decision mechanisms superior. The critical observations to consider when improving new PHY solutions are listed below:
	\begin{itemize}
		\item Actual systems do not operate ideally and struggle with synchronization problems, errors in feedback link, channel estimation errors, correlation, and RF front-end impairments.
		\item Wireless communication environments and numerous parameters of a communication system change dynamically and may show unexpected variations. 
		\item Analytical methods, such as the maximum a posteriori and maximum-likelihood based approaches, are employed by establishing hypotheses and assuming prior probabilities. However, these hypotheses and the corresponding probability distributions may not accurately represent realistic systems. For example, time variations or channel correlations can not be tracked accurately, and the corresponding assumptions become invalid. Furthermore, the inconsistencies in practice make it difficult to determine appropriate decision thresholds for the likelihood-based hypothesis tests. 
		\item Towards 6G, each communication system will become unique by interlacing technologies, applications, and users. Until today, these methodologies have not affected the solutions' performance apparently. However, next-generation systems will face various exceptional conditions and suffer significantly from flawed system representations. Therefore, generalized solutions with over-simplifications will not meet the targeted performance requirements. 
	\end{itemize} 
All these facts sidetrack or limit the presented solutions for PHY decision mechanisms. These limitations, arising from the assumptions, can be overcome with a changeover from the current methodologies to learning-driven methodologies.
\subsection{Benefits of Learning-driven Methodologies}
Learning-driven methodologies have a promising future for PHY decision mechanisms as well as in various parts of communication systems due to their attractive benefits. The inclusion of learning-driven methodologies in future solutions is motivated by the following benefits:
	\begin{itemize}
		\item The aforementioned problems due to confined system representations can be eliminated by providing data based on actual systems.
		\item A unique PHY decision mechanism can be built thanks to heuristic learning ability. If an inclusive dataset is available, learning algorithms can learn and decide based on the inherent characteristics of the system. 
		\item The decision mechanism based on learning-driven methodologies can be capable of self-evolving and continuously improved. These capabilities make it more flexible and compatible with actual conditions.
		\item Autonomy can be increased by individual PHY decisions without human interference.
	\end{itemize}
These motivations corroborate the idea that learning-driven methodologies get beyond the limits of the current methodologies. The next sections point out the main milestones and challenges to get the inference about learning-driven decision mechanisms.
\section{A Roadmap for Learning-Driven Solutions}\label{sectionIV}
\begin{figure}[!ht]
	\centering
	\includegraphics[width=0.5\textwidth]{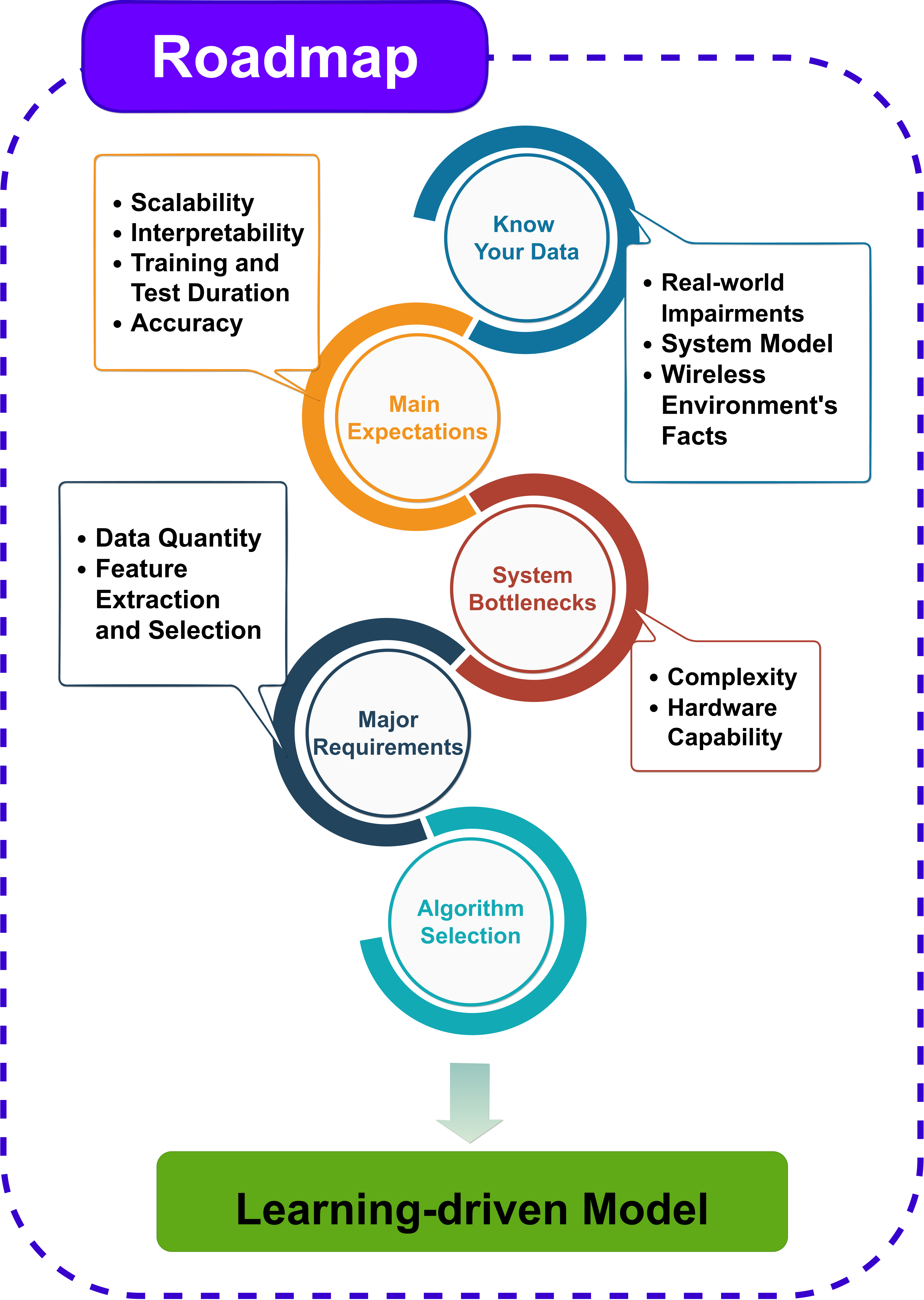}
	\caption{An illustration describing the introduction of the learning-driven solutions into PHY decision mechanisms in wireless communication systems.}
	\label{learn}
\end{figure}

Learning algorithms from shallow to deep architectures have been advancing to comprehend the system facts. However, it is unclear how an algorithm should be sifted out from several algorithms for a solution and which steps should be considered primarily. The following roadmap, as visualized in Fig. \ref{learn}, can be used to develop solutions for next-generation systems.
\renewcommand{\labelitemi}{$\blacksquare$}
\subsection{Examine the data source and know your data} {Learning algorithms do not magically provide answers to any system; they must be fed with suitable data to acquire the desired output \cite{ref16}. It can be accomplished by understanding the problem and its source. Firstly, the system and environment that render the input data to output data must be analyzed correctly. Secondly, the input/output data variations should also be carefully observed, and their relations should be considered. These steps show what is expected to learn from machine learning. For example, in a communication system, let expect machine learning to gain insights about transmitted signals via received signals. Here, the input data may typify the received signal, whereas the output data may refer to the transmitted signal. If only data based on transmitted and received signals are provided, the performance may not be sufficiently high. The reason for that the output data is formed with many factors during the transmission due to real-life impairments, and channel conditions. Then, the input data should include more information to obtain the correct output. If the knowledge about the data source exists, the input data can be enriched with more information. Additionally, it helps to find out reasons for some results during the machine learning model development stage.}
\subsection{Determine the main expectations of the proposed learning-driven model} This step has a direct impact on learning-driven solution design. Performance targets such as accuracy, interpretability, scalability, training and prediction duration are determined depending on the proposed system \cite{EXP1,EXP2,EXP3}. Some learning models, mostly based on deep learning, are not obvious in terms of model transparency and functionality. However, interpretability and low-latency of the solution are indispensable, especially for mission-critical systems. Such learning models may not be suitable in PHY, despite their accuracy. Besides the interpretability, the scalability of the models is highly crucial because wireless communication devices and their capabilities diversify widely \cite{final4,final5}. 
\subsection{Be aware of the system bottlenecks} {Employment of machine learning contains data processing, data storage, training and testing stages, and execution of generated machine learning models. All steps bring computational complexities along and require appropriate hardware capabilities. The systems may not meet high-level computational needs and constraint the solutions \cite{B1,B2,B3,B4}. Therefore, the detection of system bottlenecks is critical for the selection of the proper machine learning algorithm. The number of conceivable options for learning algorithms may decrease together with main expectations from the learning model. If the remaining algorithms do not satisfy the demand, some alternative ways should be considered to work around the system constraints. Complexity can be reduced, for example, by converting non-numeric features into numeric features or by increasing the flexibility of the system architecture.\label{systembottlenecks}}
\subsection{Detect the major requirements of the solution approach} {This milestone refers to the data quantity and quality, or major requirements for qualified models. The data quantity and quality requirements may vary depending on the results of the previous milestones' outcomes. The amount of data for the remaining algorithms through the roadmap is the key part. If a sufficient amount of data is available, model quality is dictated by data quality matters such as completeness, consistency, veracity, validity, and timeliness of the data. However, not all essential data may be readily available, especially for wireless communications, which have various destructive effects on data. Even if it is likely that the data will be collected in a timely manner, real-life impairments lead to inconsistencies, and the data will need to be processed to make it valid. Therefore, these constraints must be considered before choosing an algorithm \cite{DQ1,DQ2,DQ3}}.

\subsection{Choose and optimize the selected algorithm} {Following these instructions, the most suitable learning algorithm is chosen. It is beneficial to keep in mind the considerations of the previous steps in order to prioritize and enable the iterative process of improving the learning model. For example, if the model is overfitted, the amount of data can be increased or the complexity of the model can be reduced. If these are not possible, regularization techniques or early stopping can be considered. Afterwards, the performance of the learning model must be optimized by tuning the hyper-parameters. Here, depending on feasibility, both informed search methods and uninformed search methods such as grid search or random search can be preferred \cite{final1,final2,final3}. After training with the chosen hyper-parameter settings, the acquired learning model can be employed in the proposed system.} 
\section{Challenges for Learning-Driven Solutions}\label{sectionV}
\subsection{Data scarcity} {Data scarcity is the lack of data to generate a machine learning model. It poses a major obstacle to learning concepts based on the opinion, acquiring knowledge and insights through experiences. The main issue due to data scarcity is the overfitting of the model that leads to failure in real-time. Detecting outliers or noise in a sparse dataset is another issue. Basic solutions to avoid these problems are to favor simpler or linear techniques, and feature engineering. There are also further techniques for different data scarcity problems that can be divided into five categories: no data, rare data, small data, unlabeled data, and imbalanced data \cite{D1,D2,D3,D4,D5,D6,D7,D8,D9}. Here, no data, the absence of any data is the toughest but a possible condition due to privacy, security, and confidentiality concerns. However, if the problem is clearly defined and structured, several options can serve the purpose such as open-source datasets, encrypting or anonymizing data, and federated or online learning. As for the other scarcity problems, all techniques focus mainly on the following goals: increasing the amount of data or reducing the need for big data.
\subsubsection{Increasing the amount of data} 
Increasing the amount of data can be considered for small data, imbalanced data, and unlabeled data. A well-known solution to enlarge datasets is data augmentation, which is the acquisition of new data by diversifying existing data such as cropping, random insertion, modifying, and transforming. Synthetic data generation can be considered as another solution. Generative adversarial networks, simulation environments, and the synthetic minority over-sampling method help to form realistic data. Additionally, self-supervised learning can overcome entirely the unlabeled data challenge. When there are some labeled data but the amount is not enough, semi-supervised learning and weak-supervision should be considered.
\subsubsection{Reducing the need for big data} 
{Reducing the need for big data can be a shortcut for small and rare data problems. Transfer learning, federated learning, and few-shot learning are prominent solutions. Transfer learning enables carrying insights based on a large dataset. On the other hand, federated learning aggregates multiple knowledge from decentralized machine learning models. The few-shot learning should be given much thought of rare datasets because it enables a generalization of less data information and prior knowledge.}
\subsection{Data acquisition} {Data acquisition has a significant importance to train learning algorithms successfully. Each data sample includes the information of a certain time interval and sometimes inaccurately represents the system due to the instantaneous changes in the environment. This issue can be handled by increasing the quantity of data. However, this is not sufficient if the collected data is not qualified. Generating a qualified dataset requires overcoming the following issues:}
\begin{itemize}
	\item {The justifiability of the data-generation environment's convenience should be considered. For example, the preference of simulation-based datasets in a communication system can not be satisfactory to prove the learning-driven solution's performance due to the differences between simulations and the real-world. Real datasets can be provided at least as an explicit reference to confirm the consistency between the datasets from different sources.}
	\item {The generation of the accurate features included in the dataset is a crucial issue. Features should consist of the information individually or be jointly related to the problem and its solution. For example, features can be created depending on CSI to express estimation errors and correlation problems in PHY due to the impacts of channel conditions, as in \cite{ref12}.}
	\item {The correctness of the dataset can not be verified easily due to the lack of standardization for the labeling procedure.}
	\item {The trade-off between the size of data and the model performance should be addressed. Each additional feature expands the dataset dimensionally and leads to an increase in the training and prediction complexity. However, feature extraction or selection methods can be employed to overcome the problem, as described below.}
\end{itemize}
\subsection{Feature extraction or selection} {The extraction and selection of features are the main concepts to leverage the performance of learning algorithms, and both of them improve data representation. The difference between them is the fact that feature extraction is more general and proposes to create useful features by utilizing the existing data while the feature selection targets increasing the relevancy by removing redundant features \cite{ref16}. The selection of features provides a clarified representation of the data in lower-dimensional space by filtering. Therefore it is a highly effective approach, especially for problems that have a computationally intensive dataset. For instance, the analysis of the correlation and mutual information between the features, the determination of rating benchmark, and building a score function by weighting features are applied to detect the optimal feature subsets. The derivation/extraction of new features can be realized with linear and non-linear transformation methods such as principal component analysis, linear discriminant analysis, and autoencoder.}

\subsection{Computational complexity} {A learning process consists of two main stages: model training and prediction. Both bring a computational complexity to the system. The complexities of these processes vary due to several factors such as the selected learning algorithm and the parameter values for the model architecture's tuning and design. In the recent years, deep learning techniques have shown tremendous successes in broad application areas, and the trade-off between choosing less complex algorithms and attaining a higher accuracy is observed. The training complexity is overcome by the transfer learning, and the prediction complexity can be reduced with computational offloading or collaborative techniques \cite{C1,C2,C3}}.
\subsection{Hardware capabilities} {Learning algorithms require hardware competence for data pre-processing, model training and testing. The utilization of advanced methods such as deep or ensemble learning techniques requires large quantities of the following resources: processing power, physical size, cost, and memory. The hardware should include sufficient memory to store variables, dataset, and the trained model besides computational power. At this point, the implementation of learning algorithms on computationally constrained devices becomes a complicated problem, especially in PHY. For example, many edge Internet of Things (IoT) devices in the industry are insufficient to realize the training process or store the dataset. Cloud platforms are a possible solution to store data, build models, or control the devices remotely. However, cloud-based solutions entail sturdy communication between the user and the cloud. This introduces another load in communication systems.}
\subsection{Security} {The significant challenge is meeting the security necessities of data and the learning model to centralize the learning algorithm as a solution. Whilst robustness of the model is provided against the model's replicas, defense mechanisms should be investigated for data preservation and user privacy. Poisoning or evasion attacks threaten the data integrity and damage the tenacity of the model besides continuous changes in the channel and attack types. At this point, security maintenance and management for learning-driven solutions become more critical issues \cite{sec1,sec2,sec3}. However, the capabilities of learning algorithms offer advantages to enhance the security against unpredictable dynamics and various attacks, including modification, denial-of-service, malware, and message replay.}
\begin{figure}[!hb]
	\centering
	\includegraphics[width=0.5\textwidth]{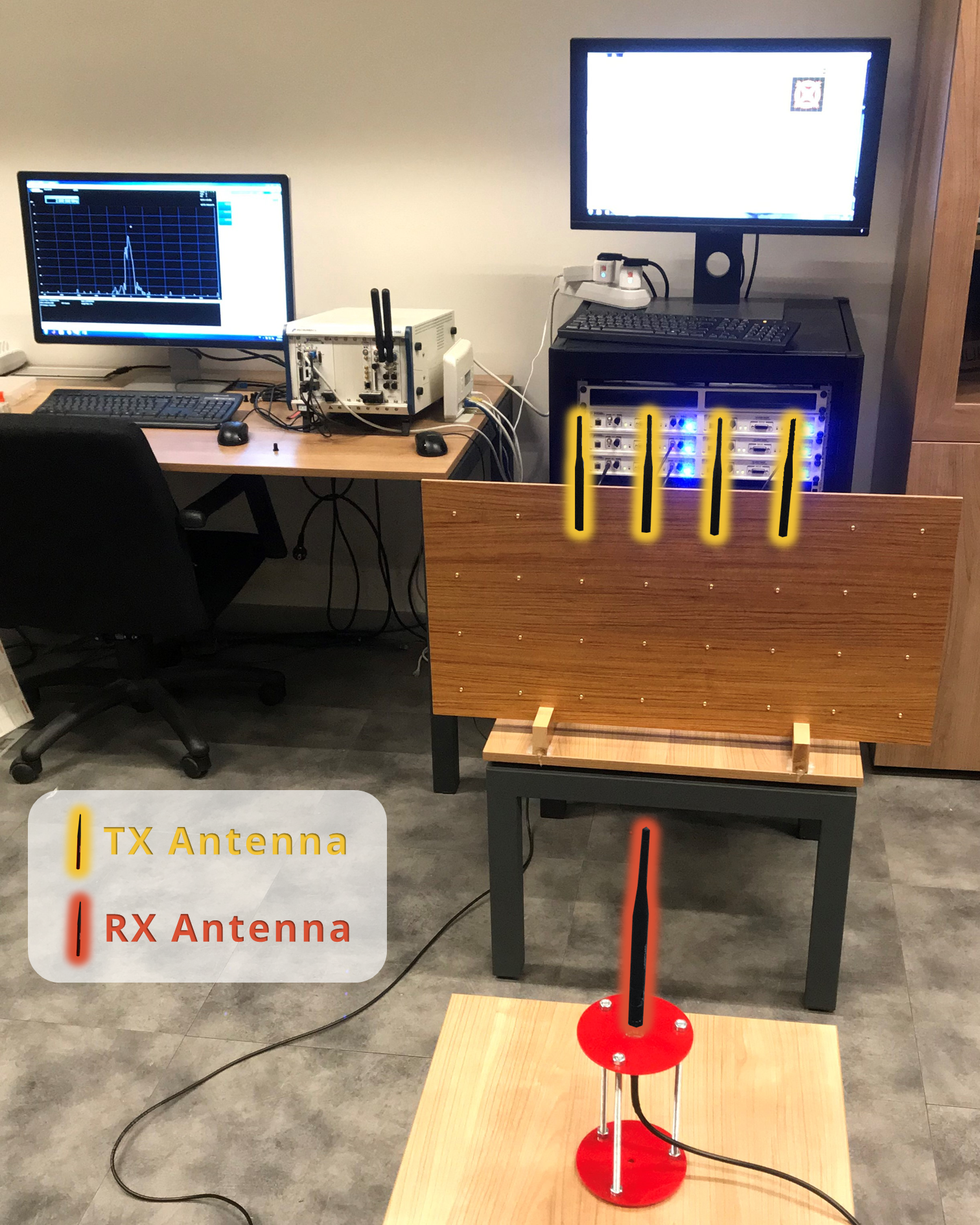}
	\caption{$4\times1$ test-bed which performs antenna selection at the transmitter.}
	\label{testbed}
\end{figure}
\begin{figure*}[!ht]
	\begin{subfigure}{0.5\textwidth}
		\centering
		\includegraphics[width=\textwidth]{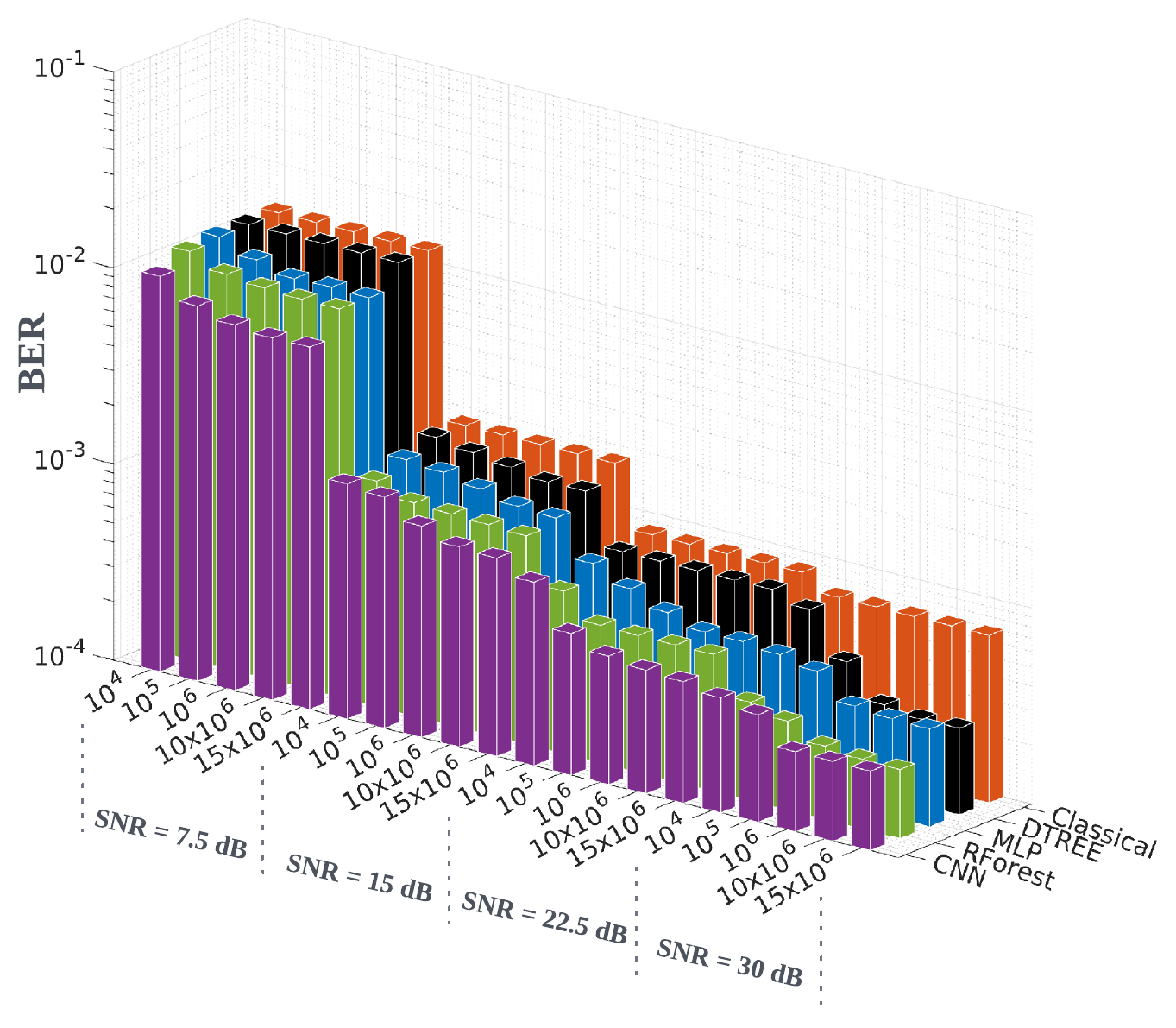}
		\caption{}
		\label{fig:8bit}
	\end{subfigure}
	\begin{subfigure}{0.5\textwidth}
		\centering
		\includegraphics[width=\textwidth]{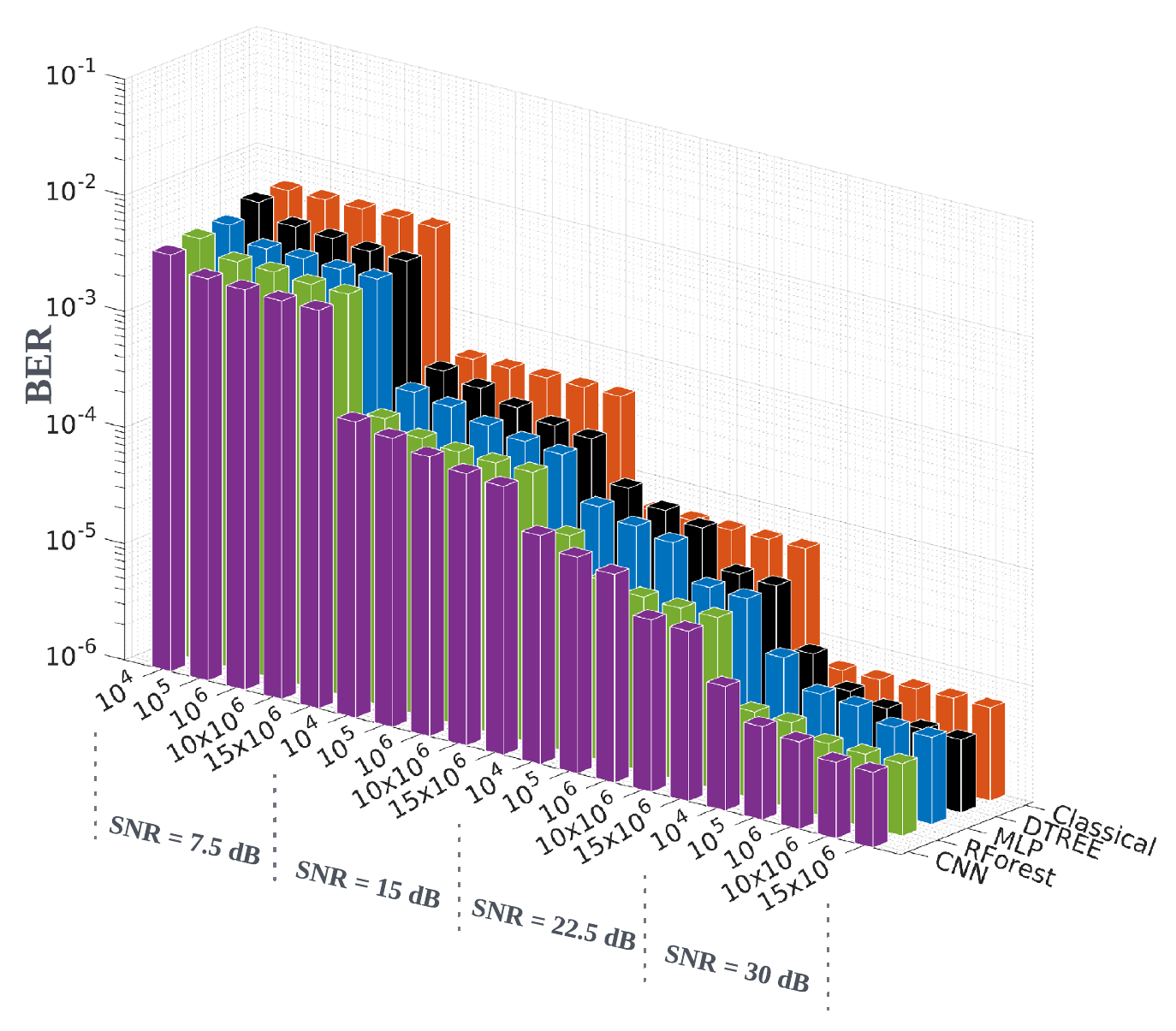}
		\caption{}
		\label{fig:16bit}
	\end{subfigure}
	\caption{The comparative results of learning algorithms and maximum-likelihood (classical) based detector. a), and b) The results for 8-bit and 16-bit ADC/DAC, respectively.}
	\label{fig:results}
\end{figure*}

\section{Software-Defined Radio Based Case Study}\label{sectionVI}
We aim to show the capabilities of learning-driven approaches on real systems without theoretical approaches' common assumptions through a benchmark. In this section, the following considerations discussed in previous sections are highlighted with the proposed case study;
\begin{itemize}
	\item A real system must overcome the front-end impairments of RF transceivers to sustain the efficacy such as IQ imbalance distortions or phase noise \cite{ref13}, (\textit{issues about synchronization, RF front-end impairments}).
	 
	\item The increase in the number of antennas brings new difficulties \cite{ref14}. For example, obtaining perfect CSI is not possible at the transmitter, (\textit{issues about channel estimation errors, correlation, and RF front-end impairments}).
	
	\item The multiple antennas require an larger number of RF chains, ADC/DAC and larger volatile memory. However, the rise in the number of converters leads to the emergence of enormous data stacks in IQ planes, and the management of these data is another challenge. They can be minimized by reducing the bit-resolution of the conversion, but then quantization errors must be taken into account. The trade-off between the converter resolution and quantization errors is the another challenge, (\textit{issues about hardware constraints, and RF front-end impairments}).
	
	\item Machine learning algorithms have different accuracy performances and complexities. Even though, we expect high performance which may require a complex architecture, some edge devices may not have suitable hardware, (\textit{issues about hardware constraints, computational complexity, and accuracy performance}).
	
	\item Transmission conditions in a wireless environment may show variations due to interference, mobility, or noise, (\textit{issues about data quantity, and feature extraction or selection}).
	
	\item The impact of the data amount on performance is important, especially some learning algorithms are more vulnerable. However, there may not be enough data available for the training process, (\textit{issue about data scarcity and accuracy performance}).
	
	\item Even if data scarcity is not a problem, the time-complexity of the training process and the required storage space increase with larger data sets, (\textit{issues about hardware constraints and computational complexity}). 
	
\end{itemize}
We proposed a learning-driven solution for antenna selection decision that is performed at the transmitter with the goal of reducing the bit error rate (BER). Antenna selection decision performance is separately investigated through 200 different cases (2 ADC/DAC resolutions, 4 SNR values, 5 training data amount, and 5 algorithms) with a test-bed design by considering aforementioned issues. Measurements are taken from a 4$\times$1 multiple-input-single-output test-bed, which is constructed using SDR units. The measurement-based performance results allow us to observe the composite impact of these challenges in the BER results, as seen in Figure \ref{fig:results}.
 
\subsection{Test-Bed Design}
The dataset is prepared via the test-bed demonstrated in Fig. \ref{testbed}. It is designed with five SDR units, USRP-2943Rs at the transmitter and receiver. Each one is used with two RF chains. To provide hardware synchronization: reference clock is generated and shared via all USRP-2943Rs by using CDA-2990 $8$ channel clock distribution accessory. The transmitter's operating frequency and bandwidth are tuned as $2.45$ GHz and $1$ MHz. As a single-carrier modulation method, BPSK is used, with a root-raised cosine filter of roll-off factor $0.5$. The distance between the transmitter and the receiver is set to $1.5$ meters. $32$ symbols are used for the acquisition, and the data/pilot rate is selected as $5/1$ at an IQ rate of $125$ ksample/s. CSI feedback is obtained via time-division duplex feedback.

\subsection{Machine Learning}
The proposed issue is defined as a classification problem. Four machine learning and deep learning algorithms are chosen: DTREE, MLP, RForest, CNN. The selection of algorithms is realized by considering the system constraints and expectations from learning models. The RForest's depth, DTREE's depth, MLP's hidden layer number, and neurons at one layer are set as $25$, $15$, $2$, $10$, respectively. As a deep learning algorithm, 1D-CNN is chosen to present a fair comparison. The CNN architecture is designed with a one-dimensional convolution layer ($256$ filters) and two fully-connected layers ($128$ and $6$ neurons, respectively). The batch size of CNN and MLP is selected as $1024$. Training of the CNN is performed by utilizing the categorical-cross entropy loss function and the adaptive moment estimation function. The learning rate $0.001$ and the exponential decay rates for the first and second moment estimates $0.99$ and $0.999$ are set.

\subsection{Results}
\begin{figure*}[!ht]
	\centering
	\includegraphics[width=\textwidth]{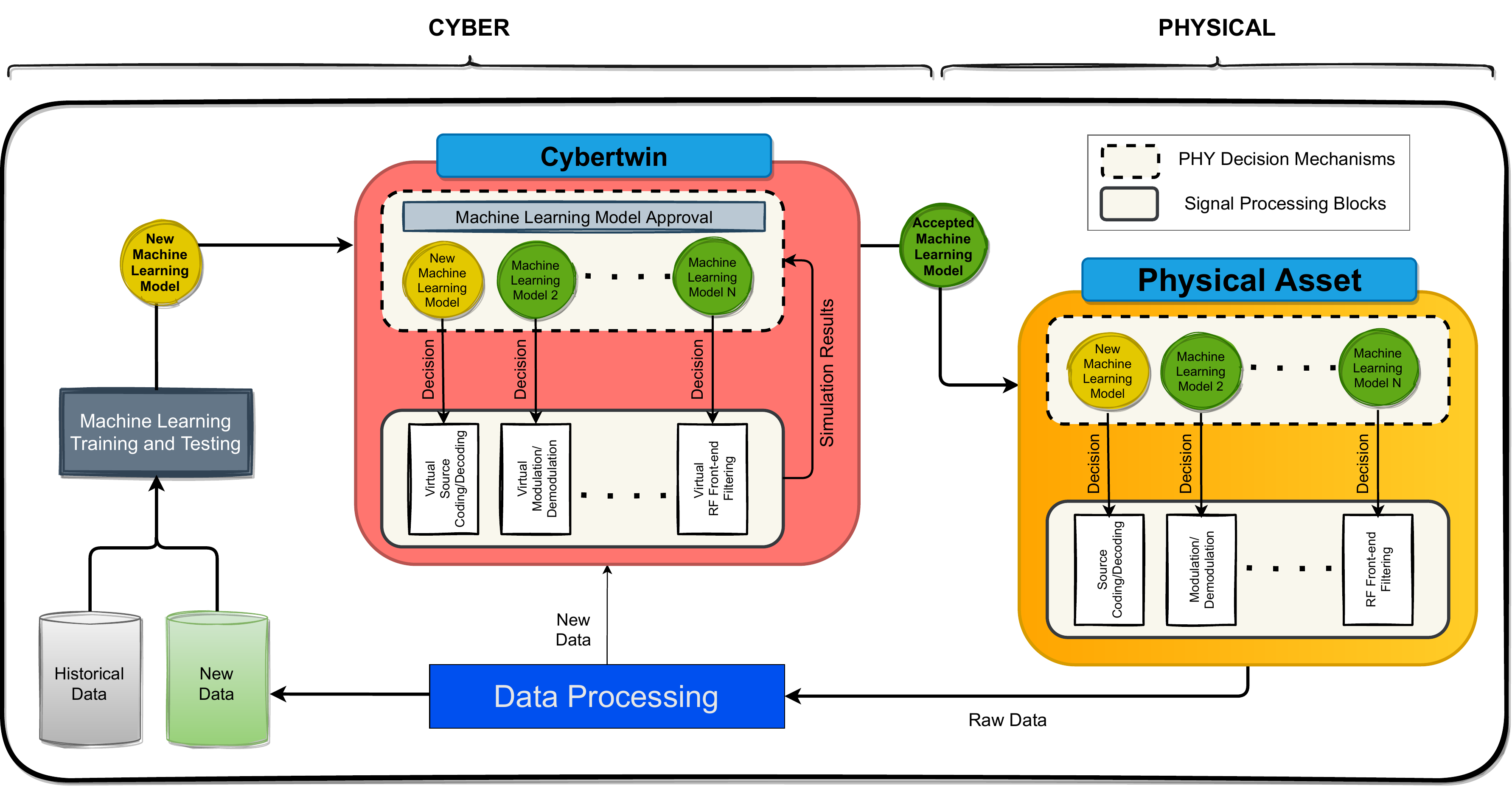}
	\caption{An intelligent cyber-physical framework for future remedies of next-generation communication systems.}
	\label{cybertwin}
\end{figure*}
The performances of RForest, DTREE, CNN, and MLP are compared with the conventional detector based on maximum-likelihood in terms of the BER measurements. Least-square-based channel estimation is performed in maximum-likelihood along with Moose's algorithm adopted to a single carrier as described in \cite{CFO}. A comparative illustration of their capabilities against the aforementioned practical challenges is shown in Fig. \ref{fig:results}. The algorithms are trained with an offline learning mechanism by performing the same analysis for the different training data sizes. When the data size becomes larger, the learning algorithms achieve a higher performance than the conventional method because of two reasons. The first reason is that numerous data samples carry the pervasive pattern of the system and represent conceivably. The second reason is the ability of heuristic learning. Moreover, these results prove the importance of data quantity and the performance flexibility of learning algorithms. Even though the necessary increase in data amount is not remarkable to reach a higher performance in this study, note that the required data amount depending on the problem can change to provide a desired leap of performance.

We can see that CNN and RForest outperform the other, less complex methods, while} the conventional method generally shows a lower performance for each SNR and converter resolution. Although CNN performs better than RForest at lower SNR values for the 8-bit ADC/DAC resolution test-bed, RForest improves performance for the 16-bit ADC/DAC resolution test-bed. MLP and DTREE offer mediocre performance compared to their learning-driven and conventional counterparts. However, if the computational limitations are considered, their performances are substantially preferable. Furthermore, DTREE, after rising steeply, surpasses CNN and MLP and approaches RForest in the case of the 16-bit converter at a high SNR. Even if it offers a simpler structure, it does not provide the capability to capture non-linear relationships between data. Therefore, its performance is closer to maximum-likelihod detector than neural network based algorithms. 

The decrease in the converters' resolution leads to an explicit increment in BER, as shown in Figs. \ref{fig:8bit} and \ref{fig:16bit}. BER differences increase, especially in high SNR. Although 16-bit ADC/DAC utilization is reasonable to sustain the reliability, this means two times IQ data, memory space, and complexity according to the number of antennas. The enormous data stacks in the IQ plane can be reduced by including feature extraction. The results show that feature extraction is a favorable process to defeat the hardware constraints in PHY and take advantage of the reliability besides boosting the learning algorithms’ performance.\\

\section{Future Remedies In An Embracing Framework}\label{sectionVII}

{The presented SDR-based case study substantiates the influences of the oversimplifying assumptions in Section \ref{sectionII} and the challenges in Section \ref{sectionV}. Machine learning can boost the system by eliminating oversimplifying assumptions and including reasoning. However, it does not become the sole remedy to meet the requirements of next-generation systems that are the compound of various cyber and physical processes. Considering these, several remedial solutions and concepts expand into the literature besides machine learning such as integrated sensing and communications \cite{remedy3,remedy4,remedy7}, IoT, distributed ledgers \cite{remedy1, remedy2}, edge-fog-cloud computing \cite{remedy5,remedy6,remedy12}, and signal processing techniques \cite{remedy8,remedy9,remedy10,remedy11, remedy13, remedy18,9793699}. In this section, we aim to provide an embracing framework for future remedies at cyber and physical levels.}

{The aforementioned and much more remedies have led up to an innovative view: cyber-physical systems (CPS) that are found on five fundamental aspects; sensing, communication, computing, control (monitoring and reasoning), and actuation. In the literature, CPS is mostly associated with IoT-based solutions for smart cities, Industry 4.0, or vehicular technologies \cite{cyber1, cyber2, cyber3, cyber4, cyber5, cyber6, cyber7, cyber8}. Beyond these solutions, communication systems at all scales and levels need to be formed into a cyber-physical framework to take advantage of upcoming remedies. They should also be equipped with learning-driven decision mechanisms and semantically upgraded. Therefore, we propose an intelligent and embracing cyber-physical framework for communication systems as demonstrated in Fig. \ref{cybertwin}. The proposed framework consists of three main operational parts: physical asset, machine learning, and cybertwin.}

\subsection{Physical Asset}
{The physical asset represents a tangible "thing" of the real-world communication systems. From the PHY perspective, it includes consecutive decision steps related to various transmission parameters such as coding rate, modulation type, constellation size, number of active antennas, transmit power, CFO, TO, channel estimation errors, or correlations. All these parameters feed signal processing blocks and must be adaptable to the variable conditions of the actual systems. Therefore, decision mechanisms are formed as machine learning models to provide more rational decisions. Each signal processing block is configured depending on the decision of the associated machine learning model, which is computed and controlled using historical running data. Here, secure data migration from the physical asset still needs to be investigated for different scenarios of future systems where the physical asset will not only be subject to different constraints but also to diverse set of applications.}

\subsection{Machine Learning}
{The deep development part of the proposed framework compromises data processing and storage, as well as machine learning training and testing. Thanks to numerous advances, this part of the framework can be flexibly designed for future communication scenarios. The roadmap, as explained in Section \ref{sectionIII}, sets the focus of these operations, such as running fast algorithms, making approximate or exact predictions, and tolerance to dynamism or time lags in the data stream. A lot of remedial solutions from training strategies to state-of-the-art learning methods can be executed depending on the focus \cite{remedy19, remedy20,remedy21,remedy22,remedy23,remedy24}. For example, when the focus is on accuracy, data processing may be more sophisticated and deep learning algorithms may be preferred. If avoiding the time lags in the data stream is the criterion, machine learning computations may take place at the edge rather than in the fog or in the cloud. Differently, federated learning can be applied as a training strategy to strengthen the dynamism, or transfer learning for latency vulnerability. However, data compression and reconstruction techniques still need to be explored to facilitate machine learning part for different scenarios and massive data sets. Although machine learning techniques can be equipped with numerous advances in the design phase, they do not guarantee that the last model generated will be better than the previous one. Future research studies need to focus on the robustness of learning models against model instability \cite{final6}. In addition, there is a mandatory examination before the model and the physical asset interaction but accomplishing this with human control seems unfeasible. Automated monitoring and developmental control of the model and the physical asset are crucial.}

\subsection{Cybertwin}
{Cybertwin is a bridge between the physical asset and machine learning, and a virtual representation of the physical asset in cyberspace. It provides automated monitoring, control, and improvement by enabling simulations through the data \cite{ctwin1,ctwin2,ctwin3}. Cybertwin is continuously updated with a new machine learning model and real data. If a new machine learning model meets the acceptance criteria after the simulations, it is implemented on the physical asset. It also checks the association of multiple machine learning models that are responsible for different signal processing blocks. These capabilities improve the flexibility, fidelity, and reliability of machine learning-driven decision mechanisms and leverage the overall performance. However, a lightweight and secure communication protocol for interaction between the cybertwin and other operational parts should be improved to take advantage of cybertwin \cite{final7}.}

\section{Conclusion}\label{sectionVIII}
This article presents a comprehensive overview of the wireless communication systems' prominent conjectures in PHY and how they may introduce modeling errors in real-world systems. From this aspect, learning algorithms are addressed with an elucidating guideline to defeat their classical counterparts and eliminate the need for oversimplifying and impractical assumptions. Additionally, the challenges to employ learning-driven solutions in PHY are presented for a holistic view. The listed considerations are supported by a real-time SDR-based case study.
The results prove that learning algorithms will take a role as a key technology for future avenues of wireless communication. At this stage, future studies should focus on how learning algorithms can be implemented with other technologies. In the last section, a holistic cyber-physical framework is proposed for joint implementation with future remedies. Another direction of future studies will be the challenges of learning-driven solutions, especially trustworthiness and interpretability \cite{conclusion}.

\bibliographystyle{IEEEtran}
\bibliography{final_submission}

\begin{thebibliography}{100}
\providecommand{\url}[1]{#1}
\csname url@samestyle\endcsname
\providecommand{\newblock}{\relax}
\providecommand{\bibinfo}[2]{#2}
\providecommand{\BIBentrySTDinterwordspacing}{\spaceskip=0pt\relax}
\providecommand{\BIBentryALTinterwordstretchfactor}{4}
\providecommand{\BIBentryALTinterwordspacing}{\spaceskip=\fontdimen2\font plus
\BIBentryALTinterwordstretchfactor\fontdimen3\font minus
  \fontdimen4\font\relax}
\providecommand{\BIBforeignlanguage}[2]{{%
\expandafter\ifx\csname l@#1\endcsname\relax
\typeout{** WARNING: IEEEtran.bst: No hyphenation pattern has been}%
\typeout{** loaded for the language `#1'. Using the pattern for}%
\typeout{** the default language instead.}%
\else
\language=\csname l@#1\endcsname
\fi
#2}}
\providecommand{\BIBdecl}{\relax}
\BIBdecl

\bibitem{int1}
L.~Bariah, L.~Mohjazi, S.~Muhaidat, P.~C. Sofotasios, G.~K. Kurt,
  H.~Yanikomeroglu, and O.~A. Dobre, ``A prospective look: Key enabling
  technologies, applications and open research topics in {6G} networks,''
  \emph{IEEE Access}, vol.~8, pp. 174\,792--174\,820, 2020.

\bibitem{int2}
Z.~Zhang, Y.~Xiao, Z.~Ma, M.~Xiao, Z.~Ding, X.~Lei, G.~K. Karagiannidis, and
  P.~Fan, ``{6G} wireless networks: Vision, requirements, architecture, and key
  technologies,'' \emph{IEEE Vehicular Technology Magazine}, vol.~14, no.~3,
  pp. 28--41, 2019.

\bibitem{int3}
K.~B. Letaief, W.~Chen, Y.~Shi, J.~Zhang, and Y.-J.~A. Zhang, ``The roadmap to
  {6G}: Ai empowered wireless networks,'' \emph{IEEE Communications Magazine},
  vol.~57, no.~8, pp. 84--90, 2019.

\bibitem{int7}
G.~Jia, Z.~Yang, H.-K. Lam, J.~Shi, and M.~Shikh-Bahaei, ``Channel assignment
  in uplink wireless communication using machine learning approach,''
  \emph{IEEE Communications Letters}, vol.~24, no.~4, pp. 787--791, 2020.

\bibitem{int12}
H.~Liu, S.~Lu, M.~El-Hajjar, and L.-L. Yang, ``Machine learning assisted
  adaptive index modulation for mmwave communications,'' \emph{IEEE Open
  Journal of the Communications Society}, vol.~1, pp. 1425--1441, 2020.

\bibitem{int8}
N.~Shlezinger, N.~Farsad, Y.~C. Eldar, and A.~J. Goldsmith, ``Viterbinet: A
  deep learning based viterbi algorithm for symbol detection,'' \emph{IEEE
  Transactions on Wireless Communications}, vol.~19, no.~5, pp. 3319--3331,
  2020.

\bibitem{int9}
F.~B. Mismar, B.~L. Evans, and A.~Alkhateeb, ``Deep reinforcement learning for
  {5G} networks: Joint beamforming, power control, and interference
  coordination,'' \emph{IEEE Transactions on Communications}, vol.~68, no.~3,
  pp. 1581--1592, 2020.

\bibitem{int5}
P.~E. Gorday, N.~Erdöl, and H.~Zhuang, ``Complex-valued neural networks for
  noncoherent demodulation,'' \emph{IEEE Open Journal of the Communications
  Society}, vol.~1, pp. 217--225, 2020.

\bibitem{int6}
X.~Chen, J.~Cheng, Z.~Zhang, L.~Wu, J.~Dang, and J.~Wang, ``Data-rate driven
  transmission strategies for deep learning-based communication systems,''
  \emph{IEEE Transactions on Communications}, vol.~68, no.~4, pp. 2129--2142,
  2020.

\bibitem{ref12}
S.~{Gecgel}, C.~{Goztepe}, and G.~{Karabulut Kurt}, ``Transmit antenna
  selection for large-scale {MIMO} {GSM} with machine learning,'' \emph{IEEE
  Wireless Commun. Lett.}, vol.~9, no.~1, pp. 113--116, Jan. 2020.

\bibitem{ref15}
S.~Gecgel, C.~Goztepe, and G.~{Karabulut Kurt}, ``Jammer detection based on
  artificial neural networks: A measurement study,'' in \emph{Proc. ACM
  Workshop Wireless Secur. and Mach. Learn.}, 2019, pp. 43--48.

\bibitem{int11}
K.~E. Kolodziej, A.~U. Cookson, and B.~T. Perry, ``{RF} canceller tuning
  acceleration using neural network machine learning for in-band full-duplex
  systems,'' \emph{IEEE Open Journal of the Communications Society}, vol.~2,
  pp. 1158--1170, 2021.

\bibitem{ref5}
H.~{Huang}, S.~{Guo}, G.~{Gui}, Z.~{Yang}, J.~{Zhang}, H.~{Sari}, and
  F.~{Adachi}, ``Deep learning for physical-layer {5G} wireless techniques:
  Opportunities, challenges and solutions,'' \emph{IEEE Wireless Commun.},
  vol.~27, no.~1, pp. 214--222, 2020.

\bibitem{ref6}
F.~{Restuccia} and T.~{Melodia}, ``Deep learning at the physical layer: System
  challenges and applications to {5G} and beyond,'' \emph{IEEE Commun. Mag.},
  vol.~58, no.~10, pp. 58--64, 2020.

\bibitem{ref18}
K.~B. {Letaief}, W.~{Chen}, Y.~{Shi}, J.~{Zhang}, and Y.~A. {Zhang}, ``The
  roadmap to {6G}: {AI} empowered wireless networks,'' \emph{IEEE Commun.
  Mag.}, vol.~57, no.~8, pp. 84--90, 2019.

\bibitem{ref7}
F.~Ling and J.~Proakis, \emph{Synchronization in Digital Communication
  Systems}.\hskip 1em plus 0.5em minus 0.4em\relax Cambridge University Press,
  2017.

\bibitem{S4}
S.~K. Sharma, T.~E. Bogale, S.~Chatzinotas, B.~Ottersten, L.~B. Le, and
  X.~Wang, ``Cognitive radio techniques under practical imperfections: A
  survey,'' \emph{IEEE Communications Surveys \& Tutorials}, vol.~17, no.~4,
  pp. 1858--1884, 2015.

\bibitem{S11}
G.~Xing, M.~Shen, and H.~Liu, ``Frequency offset and {I/Q} imbalance
  compensation for direct-conversion receivers,'' \emph{IEEE Transactions on
  Wireless Communications}, vol.~4, no.~2, pp. 673--680, 2005.

\bibitem{S1}
H.~Mercier, V.~K. Bhargava, and V.~Tarokh, ``A survey of error-correcting codes
  for channels with symbol synchronization errors,'' \emph{IEEE Communications
  Surveys {\&} Tutorials}, vol.~12, no.~1, pp. 87--96, 2010.

\bibitem{S2}
P.~Moose, ``A technique for orthogonal frequency division multiplexing
  frequency offset correction,'' \emph{IEEE Transactions on Communications},
  vol.~42, no.~10, pp. 2908--2914, 1994.

\bibitem{S3}
Z.~Zhong, P.~Chen, and T.~He, ``On-demand time synchronization with predictable
  accuracy,'' in \emph{IEEE INFOCOM}, 2011, pp. 2480--2488.

\bibitem{S5}
A.~D'Andrea, U.~Mengali, and R.~Reggiannini, ``The modified {Cramer-Rao} bound
  and its application to synchronization problems,'' \emph{IEEE Transactions on
  Communications}, vol.~42, no. 234, pp. 1391--1399, 1994.

\bibitem{S6}
J.~van~de Beek, M.~Sandell, and P.~Borjesson, ``{ML} estimation of time and
  frequency offset in {OFDM} systems,'' \emph{IEEE Transactions on Signal
  Processing}, vol.~45, no.~7, pp. 1800--1805, 1997.

\bibitem{S7}
M.~Speth, S.~Fechtel, G.~Fock, and H.~Meyr, ``Optimum receiver design for
  wireless broad-band systems using {OFDM}. {I},'' \emph{IEEE Transactions on
  Communications}, vol.~47, no.~11, pp. 1668--1677, 1999.

\bibitem{S9}
M.~Luise and R.~Reggiannini, ``Carrier frequency recovery in all-digital modems
  for burst-mode transmissions,'' \emph{IEEE Transactions on Communications},
  vol.~43, no. 2/3/4, pp. 1169--1178, 1995.

\bibitem{S12}
W.~G. Cowley and L.~P. Sabel, ``The performance of two symbol timing recovery
  algorithms for {PSK} demodulators,'' \emph{IEEE Transactions on
  Communications}, vol.~42, no.~6, pp. 2345--2355, 1994.

\bibitem{S13}
B.~Yang, K.~Letaief, R.~Cheng, and Z.~Cao, ``Timing recovery for {OFDM}
  transmission,'' \emph{IEEE Journal on Selected Areas in Communications},
  vol.~18, no.~11, pp. 2278--2291, 2000.

\bibitem{S14}
J.~Liu and J.~Li, ``Parameter estimation and error reduction for {OFDM}-based
  {WLANs},'' \emph{IEEE Transactions on Mobile Computing}, vol.~3, no.~2, pp.
  152--163, 2004.

\bibitem{S15}
H.~Meyr, M.~Moeneclaey, and S.~Fechtel, \emph{Digital Communication Receivers,
  Volume 2: Synchronization, Channel Estimation, and Signal Processing}, ser.
  Wiley Series in Telecommunications and Signal Processing.\hskip 1em plus
  0.5em minus 0.4em\relax Wiley, 1997.

\bibitem{S16}
U.~Mengali, \emph{Synchronization Techniques for Digital Receivers}, ser.
  Applications of Communications Theory.\hskip 1em plus 0.5em minus 0.4em\relax
  Springer US, 1997.

\bibitem{ref8}
A.~Nasir, S.~Durrani, R.~Kennedy, H.~Mehrpouyan, and S.~Blostein, ``Timing and
  carrier synchronization in wireless communication systems: a survey and
  classification of research in the last 5 years,'' \emph{EURASIP J. Wireless
  Commun. Network.}, vol. 2016, pp. 1--38, 2016.

\bibitem{S8}
J.~He, G.~Gu, and Z.~Wu, ``{MMSE} interference suppression in {MIMO} frequency
  selective and time-varying fading channels,'' \emph{IEEE Transactions on
  Signal Processing}, vol.~56, no.~8, pp. 3638--3651, 2008.

\bibitem{S10}
N.~Beaulieu, ``The evaluation of error probabilities for intersymbol and
  cochannel interference,'' \emph{IEEE Transactions on Communications},
  vol.~39, no.~12, pp. 1740--1749, 1991.

\bibitem{ref9}
J.~Proakis, \emph{Digital Communications}, 5th~ed.\hskip 1em plus 0.5em minus
  0.4em\relax McGraw Hill, 2007.

\bibitem{CE1}
G.~Stuber, J.~Barry, S.~McLaughlin, Y.~Li, M.~Ingram, and T.~Pratt, ``Broadband
  {MIMO}-{OFDM} wireless communications,'' \emph{Proc. IEEE}, vol.~92, no.~2,
  pp. 271--294, 2004.

\bibitem{CE2}
Y.-S. Choi, P.~Voltz, and F.~Cassara, ``On channel estimation and detection for
  multicarrier signals in fast and selective rayleigh fading channels,''
  \emph{IEEE Transactions on Communications}, vol.~49, no.~8, pp. 1375--1387,
  2001.

\bibitem{CE3}
J.-J. van~de Beek, O.~Edfors, M.~Sandell, S.~Wilson, and P.~Borjesson, ``On
  channel estimation in {OFDM} systems,'' in \emph{Veh. Technol. Conf..
  Countdown to the Wireless Twenty-First Century}, vol.~2, 1995, pp. 815--819
  vol.2.

\bibitem{CE4}
M.~Medard, ``The effect upon channel capacity in wireless communications of
  perfect and imperfect knowledge of the channel,'' \emph{IEEE Transactions on
  Information Theory}, vol.~46, no.~3, pp. 933--946, 2000.

\bibitem{CE5}
Y.~Liao, G.~Sun, Z.~Cai, X.~Shen, and Z.~Huang, ``Nonlinear kalman filter-based
  robust channel estimation for high mobility {OFDM} systems,'' \emph{IEEE
  Transactions on Intelligent Transportation Systems}, vol.~22, no.~11, pp.
  7219--7231, 2021.

\bibitem{CE6}
A.~A. Khuwaja, Y.~Chen, N.~Zhao, M.-S. Alouini, and P.~Dobbins, ``A survey of
  channel modeling for {UAV} communications,'' \emph{IEEE Communications
  Surveys \& Tutorials}, vol.~20, no.~4, pp. 2804--2821, 2018.

\bibitem{F1}
H.~Zhang, S.~Wei, G.~Ananthaswamy, and D.~L. Goeckel, ``Adaptive signaling
  based on statistical characterizations of outdated feedback in wireless
  communications,'' \emph{Proc. IEEE}, vol.~95, no.~12, pp. 2337--2353, 2007.

\bibitem{F2}
Q.~Ma and C.~Tepedelenlioglu, ``Practical multiuser diversity with outdated
  channel feedback,'' \emph{IEEE Transactions on Vehicular Technology},
  vol.~54, no.~4, pp. 1334--1345, 2005.

\bibitem{F4}
H.~A. Suraweera, M.~Soysa, C.~Tellambura, and H.~K. Garg, ``Performance
  analysis of partial relay selection with feedback delay,'' \emph{IEEE Signal
  Processing Letters}, vol.~17, no.~6, pp. 531--534, 2010.

\bibitem{F5}
Y.~Teng, M.~Liu, and M.~Song, ``Effect of outdated {CSI} on handover decisions
  in dense networks,'' \emph{IEEE Communications Letters}, vol.~21, no.~10, pp.
  2238--2241, 2017.

\bibitem{F3}
C.~Jonietz, W.~H. Gerstacker, and R.~Schober, ``Robust transmit processing for
  frequency-selective fading channels with imperfect channel feedback,''
  \emph{IEEE Transactions on Wireless Communications}, vol.~7, no.~12, pp.
  5356--5368, 2008.

\bibitem{F6}
Y.~Shi, M.~Badi, D.~Rajan, and J.~Camp, ``Channel reciprocity analysis and
  feedback mechanism design for mobile beamforming systems,'' \emph{IEEE
  Transactions on Vehicular Technology}, vol.~70, no.~6, pp. 6029--6043, 2021.

\bibitem{IQ1}
A.~Tarighat, R.~Bagheri, and A.~Sayed, ``Compensation schemes and performance
  analysis of {IQ} imbalances in {OFDM} receivers,'' \emph{IEEE Transactions on
  Signal Processing}, vol.~53, no.~8, pp. 3257--3268, 2005.

\bibitem{IQ2}
N.~Kolomvakis, M.~Matthaiou, and M.~Coldrey, ``{IQ} imbalance in multiuser
  systems: Channel estimation and compensation,'' \emph{IEEE Transactions on
  Communications}, vol.~64, no.~7, pp. 3039--3051, 2016.

\bibitem{ref10}
T.~Schenk, \emph{{RF} Imperfections in High-Rate Wireless Systems: Impact and
  Digital Compensation}.\hskip 1em plus 0.5em minus 0.4em\relax Springer, 2008.

\bibitem{ref1}
S.~M. Kay, \emph{Fundamentals of Statistical Signal Processing: Estimation
  Theory}.\hskip 1em plus 0.5em minus 0.4em\relax Prentice Hall, 1997.

\bibitem{ref16}
K.~P. Murphy, \emph{Machine Learning: A Probabilistic Perspective}.\hskip 1em
  plus 0.5em minus 0.4em\relax MIT Press, 2012.

\bibitem{EXP1}
S.~Sinha, P.~Franciosa, and D.~Ceglarek, ``Building a scalable and
  interpretable bayesian deep learning framework for quality control of free
  form surfaces,'' \emph{IEEE Access}, vol.~9, pp. 50\,188--50\,208, 2021.

\bibitem{EXP2}
K.~Chen, Q.~Kong, Y.~Dai, Y.~Xu, F.~Yin, L.~Xu, and S.~Cui, ``Recent advances
  in data-driven wireless communication using {G}aussian processes: A
  comprehensive survey,'' \emph{China Communications}, vol.~19, no.~1, pp.
  218--237, 2022.

\bibitem{EXP3}
Z.~Ye, K.~Wang, Y.~Chen, X.~Jiang, and G.~Song, ``Multi-{UAV} navigation for
  partially observable communication coverage by graph reinforcement
  learning,'' \emph{IEEE Transactions on Mobile Computing}, pp. 1--1, 2022.

\bibitem{final4}
Y.~Xu, F.~Yin, W.~Xu, C.-H. Lee, J.~Lin, and S.~Cui, ``Scalable learning
  paradigms for data-driven wireless communication,'' \emph{IEEE Communications
  Magazine}, vol.~58, no.~10, pp. 81--87, 2020.

\bibitem{final5}
L.~U. Khan, W.~Saad, Z.~Han, E.~Hossain, and C.~S. Hong, ``Federated learning
  for internet of things: Recent advances, taxonomy, and open challenges,''
  \emph{IEEE Communications Surveys \& Tutorials}, vol.~23, no.~3, pp.
  1759--1799, 2021.

\bibitem{B1}
J.~Borrego-Carazo, D.~Castells-Rufas, E.~Biempica, and J.~Carrabina,
  ``Resource-constrained machine learning for {ADAS}: A systematic review,''
  \emph{IEEE Access}, vol.~8, pp. 40\,573--40\,598, 2020.

\bibitem{B2}
H.~Liu, W.~Ziping, H.~Zhang, B.~Li, and C.~Zhao, ``Tiny machine learning
  ({T}iny-{ML}) for efficient channel estimation and signal detection,''
  \emph{IEEE Transactions on Vehicular Technology}, 2022.

\bibitem{B3}
B.~Sudharsan, P.~Patel, J.~Breslin, M.~I. Ali, K.~Mitra, S.~Dustdar, O.~Rana,
  P.~P. Jayaraman, and R.~Ranjan, ``Toward distributed, global, deep learning
  using {IoT} devices,'' \emph{IEEE Internet Computing}, vol.~25, no.~3, pp.
  6--12, 2021.

\bibitem{B4}
S.~Gollapudi, \emph{Practical Machine Learning}.\hskip 1em plus 0.5em minus
  0.4em\relax Packt Publishing, 2016.

\bibitem{DQ1}
D.~Liu, G.~Zhu, Q.~Zeng, J.~Zhang, and K.~Huang, ``Wireless data acquisition
  for edge learning: Data-importance aware retransmission,'' \emph{IEEE
  Transactions on Wireless Communications}, vol.~20, no.~1, pp. 406--420, 2021.

\bibitem{DQ2}
Y.~Liu, X.~Guan, Y.~Peng, H.~Chen, T.~Ohtsuki, and Z.~Han, ``Blockchain-based
  task offloading for edge computing on low-quality data via distributed
  learning in the internet of energy,'' \emph{IEEE Journal on Selected Areas in
  Communications}, vol.~40, no.~2, pp. 657--676, 2022.

\bibitem{DQ3}
W.~Yu, Y.~Liu, T.~Dillon, W.~Rahayu, and F.~Mostafa, ``An integrated framework
  for health state monitoring in a smart factory employing {IoT} and big data
  techniques,'' \emph{IEEE Internet of Things Journal}, vol.~9, no.~3, pp.
  2443--2454, 2022.

\bibitem{final1}
G.~Luo, ``A review of automatic selection methods for machine learning
  algorithms and hyper-parameter values,'' \emph{Network Modeling Analysis in
  Health Informatics and Bioinformatics}, vol.~5, no.~1, pp. 1--16, 2016.

\bibitem{final2}
J.~Bergstra and Y.~Bengio, ``Random search for hyper-parameter optimization,''
  \emph{Journal of Machine Learning Research}, vol.~13, no.~2, 2012.

\bibitem{final3}
J.~Bergstra, R.~Bardenet, Y.~Bengio, and B.~K{\'e}gl, ``Algorithms for
  hyper-parameter optimization,'' \emph{Advances in Neural Information
  Processing Systems}, vol.~24, 2011.

\bibitem{D1}
V.~S. Sheng, F.~Provost, and P.~G. Ipeirotis, ``Get another label? improving
  data quality and data mining using multiple, noisy labelers,'' in \emph{ACM
  SIGKDD Int. Conf. Knowledge Discovery and Data Mining}, ser. KDD '08.\hskip
  1em plus 0.5em minus 0.4em\relax New York, NY, USA: Association for Computing
  Machinery, 2008, p. 614–622.

\bibitem{D2}
H.~He and E.~A. Garcia, ``Learning from imbalanced data,'' \emph{IEEE
  Transactions on Knowledge and Data Engineering}, vol.~21, no.~9, pp.
  1263--1284, 2009.

\bibitem{D3}
Y.~Roh, G.~Heo, and S.~E. Whang, ``A survey on data collection for machine
  learning: A big data - {AI} integration perspective,'' \emph{IEEE
  Transactions on Knowledge and Data Engineering}, vol.~33, no.~4, pp.
  1328--1347, 2021.

\bibitem{D4}
K.~Emam, L.~Mosquera, and R.~Hoptroff, \emph{Practical Synthetic Data
  Generation: Balancing Privacy and the Broad Availability of Data}.\hskip 1em
  plus 0.5em minus 0.4em\relax O'Reilly Media, Incorporated, 2020.

\bibitem{D5}
Z.~Meng, X.~Guo, Z.~Pan, D.~Sun, and S.~Liu, ``Data segmentation and
  augmentation methods based on raw data using deep neural networks approach
  for rotating machinery fault diagnosis,'' \emph{IEEE Access}, vol.~7, pp.
  79\,510--79\,522, 2019.

\bibitem{D6}
N.-T. Tran, V.-H. Tran, N.-B. Nguyen, T.-K. Nguyen, and N.-M. Cheung, ``On data
  augmentation for {GAN} training,'' \emph{IEEE Transactions on Image
  Processing}, vol.~30, pp. 1882--1897, 2021.

\bibitem{D7}
N.~V. Chawla, K.~W. Bowyer, L.~O. Hall, and W.~P. Kegelmeyer, ``{SMOTE}:
  Synthetic minority over-sampling technique,'' \emph{J. Artif. Int. Res.},
  vol.~16, no.~1, p. 321–357, jun 2002.

\bibitem{D8}
L.~Wang, M.~Han, X.~Li, N.~Zhang, and H.~Cheng, ``Review of classification
  methods on unbalanced data sets,'' \emph{IEEE Access}, vol.~9, pp.
  64\,606--64\,628, 2021.

\bibitem{D9}
C.~Geng, S.-J. Huang, and S.~Chen, ``Recent advances in open set recognition: A
  survey,'' \emph{IEEE Transactions on Pattern Analysis and Machine
  Intelligence}, vol.~43, no.~10, pp. 3614--3631, 2021.

\bibitem{C1}
M.~Khayyat, I.~A. Elgendy, A.~Muthanna, A.~S. Alshahrani, S.~Alharbi, and
  A.~Koucheryavy, ``Advanced deep learning-based computational offloading for
  multilevel vehicular edge-cloud computing networks,'' \emph{IEEE Access},
  vol.~8, pp. 137\,052--137\,062, 2020.

\bibitem{C2}
B.~Maschler and M.~Weyrich, ``Deep transfer learning for industrial automation:
  A review and discussion of new techniques for data-driven machine learning,''
  \emph{IEEE Industrial Electronics Magazine}, vol.~15, no.~2, pp. 65--75,
  2021.

\bibitem{C3}
H.~Xiao, C.~Xu, Y.~Ma, S.~Yang, L.~Zhong, and G.-M. Muntean, ``Edge
  intelligence: A computational task offloading scheme for dependent {IoT}
  application,'' \emph{IEEE Transactions on Wireless Communications}, pp. 1--1,
  2022.

\bibitem{sec1}
M.~Xue, C.~Yuan, H.~Wu, Y.~Zhang, and W.~Liu, ``Machine learning security:
  Threats, countermeasures, and evaluations,'' \emph{IEEE Access}, vol.~8, pp.
  74\,720--74\,742, 2020.

\bibitem{sec2}
Q.~Liu, P.~Li, W.~Zhao, W.~Cai, S.~Yu, and V.~C.~M. Leung, ``A survey on
  security threats and defensive techniques of machine learning: A data driven
  view,'' \emph{IEEE Access}, vol.~6, pp. 12\,103--12\,117, 2018.

\bibitem{sec3}
F.~O. Olowononi, D.~B. Rawat, and C.~Liu, ``Resilient machine learning for
  networked cyber physical systems: A survey for machine learning security to
  securing machine learning for cps,'' \emph{IEEE Communications Surveys
  Tutorials}, vol.~23, no.~1, pp. 524--552, 2021.

\bibitem{ref13}
V.~K.~V. {Gottumukkala} and H.~{Minn}, ``Capacity analysis and pilot-data power
  allocation for {MIMO}-{OFDM} with transmitter and receiver {IQ} imbalances
  and residual carrier frequency offset,'' \emph{IEEE Trans. Veh. Technol.},
  vol.~61, no.~2, pp. 553--565, 2012.

\bibitem{ref14}
L.~{Lu}, G.~Y. {Li}, A.~L. {Swindlehurst}, A.~{Ashikhmin}, and R.~{Zhang}, ``An
  overview of massive {MIMO}: Benefits and challenges,'' \emph{IEEE J. Sel.
  Topics Signal Process.}, vol.~8, no.~5, pp. 742--758, 2014.

\bibitem{CFO}
R.~Heath, \emph{Introduction to Wireless Digital Communication: A Signal
  Processing Perspective}.\hskip 1em plus 0.5em minus 0.4em\relax Prentice
  Hall, 2017.

\bibitem{remedy3}
J.~Mu, Y.~Gong, F.~Zhang, Y.~Cui, F.~Zheng, and X.~Jing, ``Integrated sensing
  and communication-enabled predictive beamforming with deep learning in
  vehicular networks,'' \emph{IEEE Communications Letters}, vol.~25, no.~10,
  pp. 3301--3304, 2021.

\bibitem{remedy4}
Y.~Cui, F.~Liu, X.~Jing, and J.~Mu, ``Integrating sensing and communications
  for ubiquitous {IoT}: Applications, trends, and challenges,'' \emph{IEEE
  Network}, vol.~35, no.~5, pp. 158--167, 2021.

\bibitem{remedy7}
D.~K. Pin~Tan, J.~He, Y.~Li, A.~Bayesteh, Y.~Chen, P.~Zhu, and W.~Tong,
  ``Integrated sensing and communication in {6G}: Motivations, use cases,
  requirements, challenges and future directions,'' in \emph{IEEE International
  Online Symposium on Joint Communications Sensing}, 2021, pp. 1--6.

\bibitem{remedy1}
L.~D. Nguyen, A.~E. Kalor, I.~Leyva-Mayorga, and P.~Popovski, ``Trusted
  wireless monitoring based on distributed ledgers over {NB-IoT}
  connectivity,'' \emph{IEEE Communications Magazine}, vol.~58, no.~6, pp.
  77--83, 2020.

\bibitem{remedy2}
P.~Danzi, A.~E. Kalor, R.~B. Sorensen, A.~K. Hagelskjaer, L.~D. Nguyen,
  C.~Stefanovic, and P.~Popovski, ``Communication aspects of the integration of
  wireless {IoT} devices with distributed ledger technology,'' \emph{IEEE
  Network}, vol.~34, no.~1, pp. 47--53, 2020.

\bibitem{remedy5}
D.~Kimovski, R.~Mathá, J.~Hammer, N.~Mehran, H.~Hellwagner, and R.~Prodan,
  ``Cloud, fog, or edge: Where to compute?'' \emph{IEEE Internet Computing},
  vol.~25, no.~4, pp. 30--36, 2021.

\bibitem{remedy6}
Y.~Zhou, L.~Tian, L.~Liu, and Y.~Qi, ``Fog computing enabled future mobile
  communication networks: A convergence of communication and computing,''
  \emph{IEEE Communications Magazine}, vol.~57, no.~5, pp. 20--27, 2019.

\bibitem{remedy12}
Q.-V. Pham, F.~Fang, V.~N. Ha, M.~J. Piran, M.~Le, L.~B. Le, W.-J. Hwang, and
  Z.~Ding, ``A survey of multi-access edge computing in {5G} and beyond:
  Fundamentals, technology integration, and state-of-the-art,'' \emph{IEEE
  Access}, vol.~8, pp. 116\,974--117\,017, 2020.

\bibitem{remedy8}
A.~I. Perez-Neira, M.~A. Vazquez, M.~B. Shankar, S.~Maleki, and S.~Chatzinotas,
  ``Signal processing for high-throughput satellites: Challenges in new
  interference-limited scenarios,'' \emph{IEEE Signal Processing Magazine},
  vol.~36, no.~4, pp. 112--131, 2019.

\bibitem{remedy9}
E.~Bjornson and P.~Giselsson, ``Two applications of deep learning in the
  physical layer of communication systems [lecture notes],'' \emph{IEEE Signal
  Processing Magazine}, vol.~37, no.~5, pp. 134--140, 2020.

\bibitem{remedy10}
F.~Rusek, D.~Persson, B.~K. Lau, E.~G. Larsson, T.~L. Marzetta, O.~Edfors, and
  F.~Tufvesson, ``Scaling up {MIMO}: Opportunities and challenges with very
  large arrays,'' \emph{IEEE Signal Processing Magazine}, vol.~30, no.~1, pp.
  40--60, 2013.

\bibitem{remedy11}
P.~Banelli, S.~Buzzi, G.~Colavolpe, A.~Modenini, F.~Rusek, and A.~Ugolini,
  ``Modulation formats and waveforms for {5G} networks: Who will be the heir of
  {OFDM?}: An overview of alternative modulation schemes for improved spectral
  efficiency,'' \emph{IEEE Signal Processing Magazine}, vol.~31, no.~6, pp.
  80--93, 2014.

\bibitem{remedy13}
L.~Dai, B.~Wang, Y.~Yuan, S.~Han, I.~Chih-lin, and Z.~Wang, ``Non-orthogonal
  multiple access for {5G}: solutions, challenges, opportunities, and future
  research trends,'' \emph{IEEE Communications Magazine}, vol.~53, no.~9, pp.
  74--81, 2015.

\bibitem{remedy18}
R.~Y. Mesleh, H.~Haas, S.~Sinanovic, C.~W. Ahn, and S.~Yun, ``Spatial
  modulation,'' \emph{IEEE Transactions on Vehicular Technology}, vol.~57,
  no.~4, pp. 2228--2241, 2008.

\bibitem{9793699}
G.~Alt{\i}n and I.~A. Arslan, ``Joint transmit and receive antenna selection
  for spatial modulation systems using deep learning,'' \emph{IEEE
  Communications Letters}, pp. 1--1, 2022.

\bibitem{cyber1}
A.~K. Tripathy, P.~K. Tripathy, A.~G. Mohapatra, N.~K. Ray, and S.~P. Mohanty,
  ``We{D}o{S}hare: A ridesharing framework in transportation cyber-physical
  system for sustainable mobility in smart cities,'' \emph{IEEE Consumer
  Electronics Magazine}, vol.~9, no.~4, pp. 41--48, 2020.

\bibitem{cyber2}
N.~Jazdi, ``Cyber physical systems in the context of industry 4.0,'' in
  \emph{IEEE Int. Conf. Automation, Quality and Testing, Robotics}, 2014, pp.
  1--4.

\bibitem{cyber3}
Z.~You and L.~Feng, ``Integration of industry 4.0 related technologies in
  construction industry: A framework of cyber-physical system,'' \emph{IEEE
  Access}, vol.~8, pp. 122\,908--122\,922, 2020.

\bibitem{cyber4}
A.~J.~C. Trappey, C.~V. Trappey, U.~H. Govindarajan, J.~J. Sun, and A.~C.
  Chuang, ``A review of technology standards and patent portfolios for enabling
  cyber-physical systems in advanced manufacturing,'' \emph{IEEE Access},
  vol.~4, pp. 7356--7382, 2016.

\bibitem{cyber5}
Y.~Feng, B.~Hu, H.~Hao, Y.~Gao, Z.~Li, and J.~Tan, ``Design of distributed
  cyber–physical systems for connected and automated vehicles with
  implementing methodologies,'' \emph{IEEE Transactions on Industrial
  Informatics}, vol.~14, no.~9, pp. 4200--4211, 2018.

\bibitem{cyber6}
F.~Amin and G.~S. Choi, ``Hotspots analysis using cyber-physical-social system
  for a smart city,'' \emph{IEEE Access}, vol.~8, pp. 122\,197--122\,209, 2020.

\bibitem{cyber7}
R.~Rodulfo, ``Smart city case study: City of coral gables leverages the
  internet of things to improve quality of life,'' \emph{IEEE Internet of
  Things Magazine}, vol.~3, no.~2, pp. 74--81, 2020.

\bibitem{cyber8}
D.~Jia, K.~Lu, J.~Wang, X.~Zhang, and X.~Shen, ``A survey on platoon-based
  vehicular cyber-physical systems,'' \emph{IEEE Communications Surveys \&
  Tutorials}, vol.~18, no.~1, pp. 263--284, 2016.

\bibitem{remedy19}
W.~Y.~B. Lim, N.~C. Luong, D.~T. Hoang, Y.~Jiao, Y.-C. Liang, Q.~Yang,
  D.~Niyato, and C.~Miao, ``Federated learning in mobile edge networks: A
  comprehensive survey,'' \emph{IEEE Communications Surveys \& Tutorials},
  vol.~22, no.~3, pp. 2031--2063, 2020.

\bibitem{remedy20}
F.~Zhuang, Z.~Qi, K.~Duan, D.~Xi, Y.~Zhu, H.~Zhu, H.~Xiong, and Q.~He, ``A
  comprehensive survey on transfer learning,'' \emph{Proc. IEEE}, vol. 109,
  no.~1, pp. 43--76, 2021.

\bibitem{remedy21}
A.~Feriani and E.~Hossain, ``Single and multi-agent deep reinforcement learning
  for ai-enabled wireless networks: A tutorial,'' \emph{IEEE Communications
  Surveys \& Tutorials}, vol.~23, no.~2, pp. 1226--1252, 2021.

\bibitem{remedy22}
C.~Zhang, P.~Patras, and H.~Haddadi, ``Deep learning in mobile and wireless
  networking: A survey,'' \emph{IEEE Communications Surveys \& Tutorials},
  vol.~21, no.~3, pp. 2224--2287, 2019.

\bibitem{remedy23}
A.~Steiner, A.~Kolesnikov, X.~Zhai, R.~Wightman, J.~Uszkoreit, and L.~Beyer,
  ``How to train your {ViT?} {D}ata, augmentation, and regularization in vision
  transformers,'' \emph{CoRR}, vol. abs/2106.10270, 2021.

\bibitem{remedy24}
S.~Jastrzebski, D.~Arpit, O.~{\AA}strand, G.~Kerg, H.~Wang, C.~Xiong,
  R.~Socher, K.~Cho, and K.~J. Geras, ``Catastrophic fisher explosion: Early
  phase fisher matrix impacts generalization,'' \emph{CoRR}, vol.
  abs/2012.14193, 2020.

\bibitem{final6}
M.~Zhang, H.~Zhang, Y.~Fang, and D.~Yuan, ``Learning-based data transmissions
  for future {6G} enabled industrial {IoT}: A data compression perspective,''
  \emph{IEEE Network}, pp. 1--7, 2022.

\bibitem{ctwin1}
S.~Shen, C.~Yu, K.~Zhang, and S.~Ci, ``Adaptive artificial intelligence for
  resource-constrained connected vehicles in cybertwin-driven {6G} network,''
  \emph{IEEE Internet of Things Journal}, vol.~8, no.~22, pp. 16\,269--16\,278,
  2021.

\bibitem{ctwin2}
Y.~Lu, S.~Maharjan, and Y.~Zhang, ``Adaptive edge association for wireless
  digital twin networks in {6G},'' \emph{IEEE Internet of Things Journal},
  vol.~8, no.~22, pp. 16\,219--16\,230, 2021.

\bibitem{ctwin3}
T.~K. Rodrigues, J.~Liu, and N.~Kato, ``Application of cybertwin for offloading
  in mobile multiaccess edge computing for {6G} networks,'' \emph{IEEE Internet
  of Things Journal}, vol.~8, no.~22, pp. 16\,231--16\,242, 2021.

\bibitem{final7}
G.~Li, C.~Lai, R.~Lu, and D.~Zheng, ``{SecCDV}: A security reference
  architecture for cybertwin-driven {6G} {V2X},'' \emph{IEEE Transactions on
  Vehicular Technology}, vol.~71, no.~5, pp. 4535--4550, 2022.

\bibitem{conclusion}
A.~Nascita, A.~Montieri, G.~Aceto, D.~Ciuonzo, V.~Persico, and A.~Pescapé,
  ``{XAI} meets mobile traffic classification: Understanding and improving
  multimodal deep learning architectures,'' \emph{IEEE Transactions on Network
  and Service Management}, vol.~18, no.~4, pp. 4225--4246, 2021.

\end{thebibliography}
\begin{IEEEbiography}[{\includegraphics[width=1in,height=1.25in,clip,keepaspectratio]{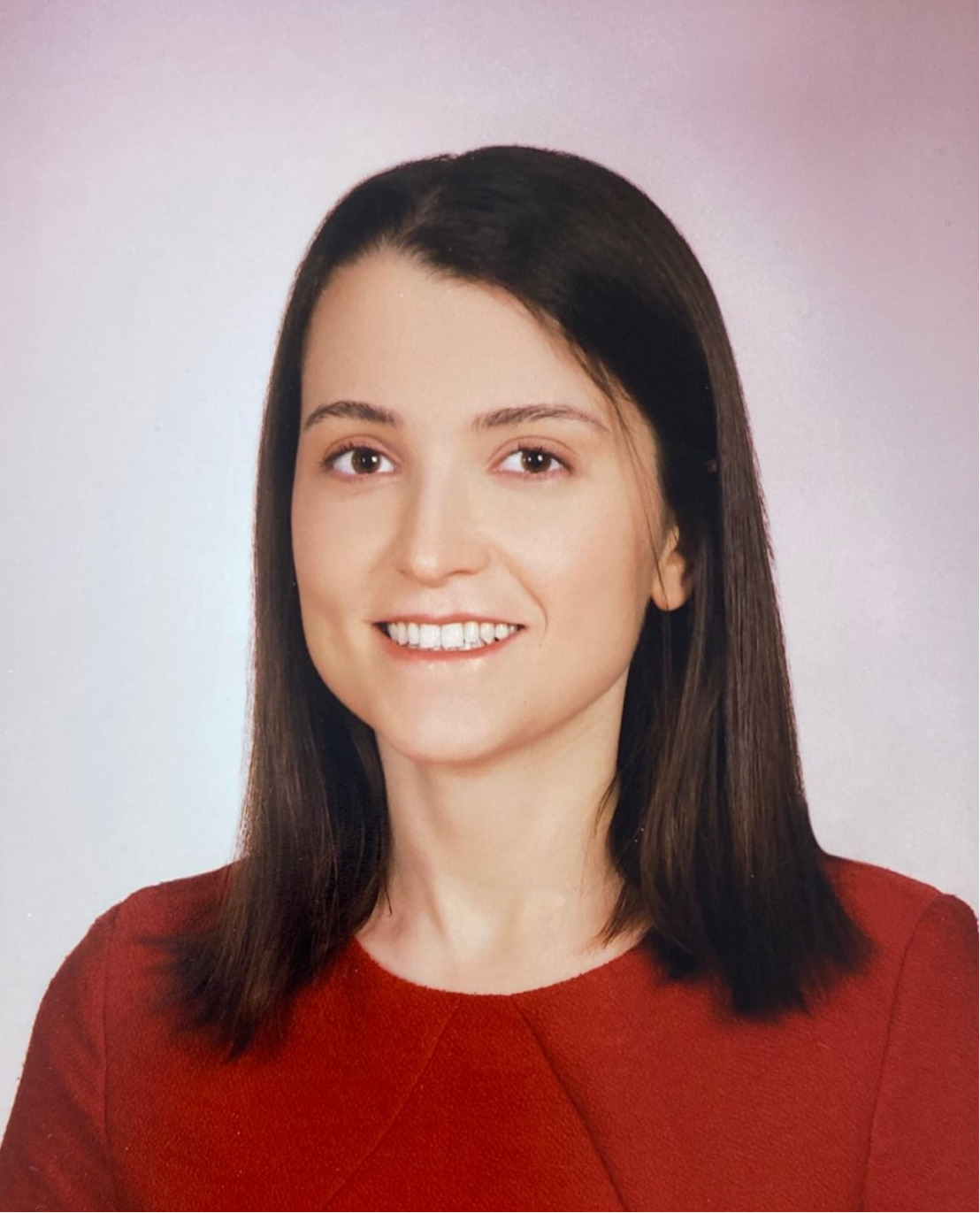}}]{SELEN GECGEL CETIN} received a B.S. degree in Electronics and Communication Engineering from Yildiz Technical University, Istanbul, Turkey, in 2016. She received an M.S. degree in Telecommunication Engineering from the Istanbul Technical University (ITU), Istanbul, Turkey, in 2019. She is currently pursuing Ph.D. degree. From 2019 to 2021, she was a Research Assistant with the Department of Electronics Engineering, Turkish Air Force Academy, NDU. She is a member of the ITU Wireless Communication Research Laboratory, and her current research is interested in machine learning and wireless communications.
\end{IEEEbiography}
\begin{IEEEbiography}[{\includegraphics[width=1in,height=1.25in,clip,keepaspectratio]{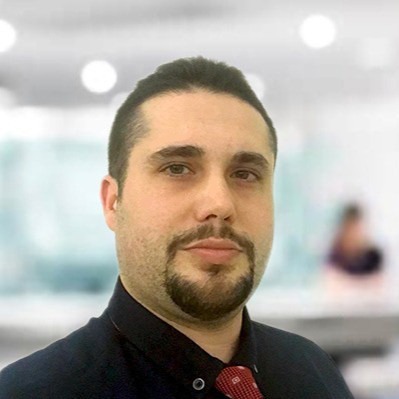}}]
{CANER GOZTEPE} received the B.Sc. and M.Sc. degrees in telecommunication engineering from Istanbul Technical University (ITU), Turkey, in 2017 and 2019. He currently manages R\&D studies on behalf of different companies in the USA, Canada, and Turkey. His main research interests are artificial intelligence, gamification, IoT, 5G+, physical layer security, and SDR applications beyond 5G physical layer schemes.
\end{IEEEbiography}
\begin{IEEEbiography}[{\includegraphics[width=1in,height=1.25in,clip,keepaspectratio]{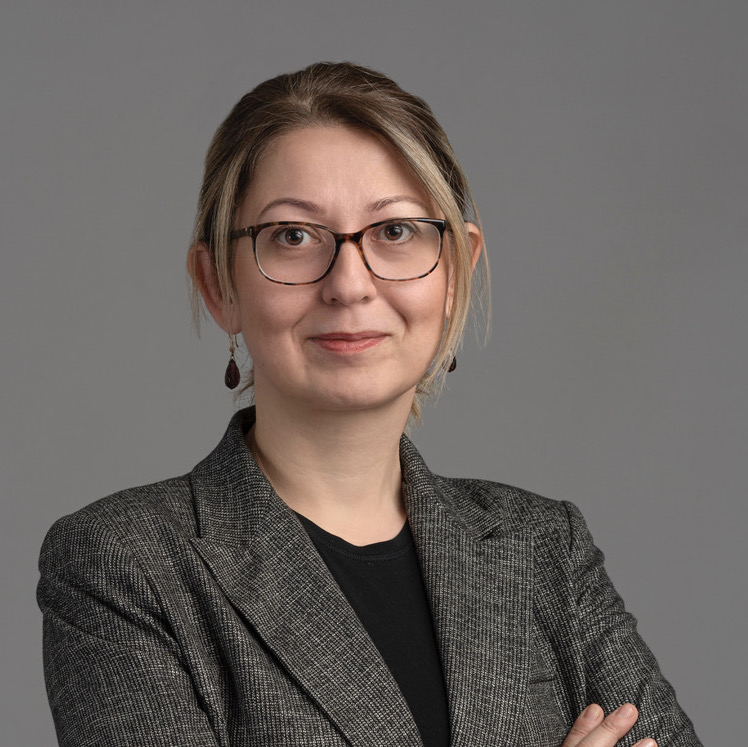}}]{GUNES KARABULUT KURT} (Senior Member, IEEE) received the B.S. degree with high honors in electronics and electrical engineering from the Bogazici University, Istanbul, Turkey, in 2000 and the M.A.Sc. and the Ph.D. degrees in electrical engineering from the University of Ottawa, ON, Canada, in 2002 and 2006, respectively. From 2000 to 2005, she was a Research Assistant with the CASP Group, University of Ottawa. Between 2005 and 2006, she was with TenXc Wireless, Canada. From 2006 to 2008, Dr. Karabulut Kurt was with Edgewater Computer Systems Inc., Canada. From 2008 to 2010, she was with Turkcell Research and Development Applied Research and Technology, Istanbul. Between 2010 and 2021, she was with Istanbul Technical University. She is currently an Associate Professor of Electrical Engineering at Polytechnique Montréal, Montreal, QC, Canada. She is a Marie Curie Fellow and has received the Turkish Academy of Sciences Outstanding Young Scientist (TÜBA-GEBIP) Award in 2019. In addition, she is an adjunct research professor at Carleton University. She is also currently serving as an Associate Technical Editor (ATE) of the IEEE Communications Magazine and a member of the IEEE WCNC Steering Board. She is the chair of the IEEE special interest group entitled “Satellite Mega-constellations: Communications and Networking”. Her current research interests include space information networks, satellite networking, wireless network coding, wireless security, space security, and wireless testbeds. 
\end{IEEEbiography}
\begin{IEEEbiography}[{\includegraphics[width=1in,height=1.25in,clip,keepaspectratio]{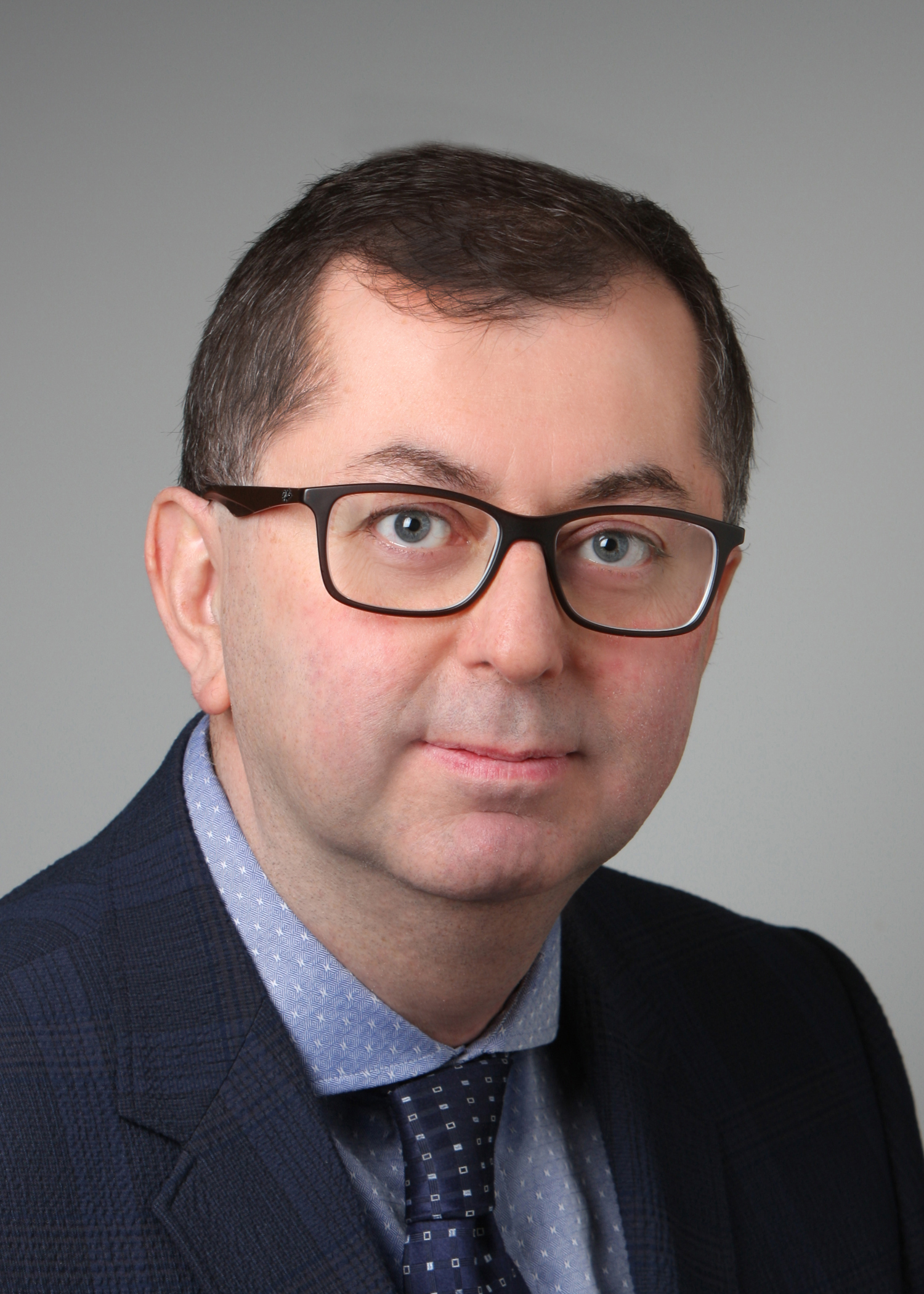}}]
{HALIM YANIKOMEROGLU} (Fellow, IEEE) is a Professor with the Department of Systems and Computer Engineering, Carleton University, Ottawa, Canada. From 2012 to 2016, he led one of the largest academic-industrial collaborative research programs on pre-standards 5G wireless. In Summer 2019, he started a new large-scale project on the 6G non-terrestrial networks. His extensive collaboration with industry resulted in 39 granted patents. He has formally supervised or hosted at Carleton a total of 135 postgraduate researchers in all levels (Ph.D. and M.A.Sc. students, PDFs, and Professors). He has coauthored IEEE papers with faculty members in 80+ universities in 25 countries and industry researchers in 10 countries. His primary research domain is wireless communications and networks. His research group has made substantial contributions to 4G and 5G wireless technologies.

Dr. Yanikomeroglu received several awards for his research, teaching, and service, including the IEEE Communications Society Wireless Communications Technical Committee Recognition Award in 2018, the IEEE Vehicular Technology Society Stuart Meyer Memorial Award in 2020, and the IEEE Communications Society Fred W. Ellersick Prize in 2021. He is currently serving as the Chair of the IEEE Wireless Communications and Networking Conference (WCNC) Steering Committee. He was the Technical Program Chair/Co-Chair of WCNC 2004 (Atlanta), WCNC 2008 (Las Vegas), and WCNC 2014 (Istanbul). He was the General Chair of IEEE VTC 2010-Fall (Ottawa) and VTC 2017-Fall (Toronto). He also served as the Chair of the IEEE’s Technical Committee on Personal Communications. He is a Fellow of Engineering Institute of Canada and Canadian Academy of Engineering and a Distinguished Speaker for both IEEE Communications Society and IEEE Vehicular Technology Society.
\end{IEEEbiography}
\end{document}